\documentclass[letterpaper]{article} 
\usepackage{aaai25}  
\usepackage{times}  
\usepackage{helvet}  
\usepackage{courier}  
\usepackage[hyphens]{url}  
\usepackage{graphicx} 
\usepackage{natbib}  
\usepackage{caption} 
\usepackage{algorithm}
\usepackage{algorithmic}
\usepackage{multirow}
\usepackage{hhline}
\usepackage{pifont}
\usepackage{booktabs}
\usepackage{amsmath}
\usepackage{amssymb}
\usepackage{graphicx}
\usepackage{xcolor}
\usepackage{enumitem}
\usepackage{colortbl}
\usepackage{newfloat}
\usepackage{listings}
\DeclareCaptionStyle{ruled}{labelfont=normalfont,labelsep=colon,strut=off} 
\floatstyle{ruled}
\newfloat{listing}{tb}{lst}{}
\floatname{listing}{Listing}
\title{MambaPro: Multi-Modal Object Re-Identification with Mamba Aggregation and Synergistic Prompt}
\author{
    Yuhao Wang\textsuperscript{\rm 1},
    Xuehu Liu\textsuperscript{\rm 2},
    Tianyu Yan\textsuperscript{\rm 1},
    Yang Liu\textsuperscript{\rm 1,4},
    Aihua Zheng\textsuperscript{\rm 3,4},\\
    Pingping Zhang\textsuperscript{\rm 1,4}\thanks{Corresponding author (zhpp@dlut.edu.cn).},
    Huchuan Lu\textsuperscript{\rm 1}
}
\affiliations{
    \textsuperscript{\rm 1}School of Future Technology, School of Artificial Intelligence, Dalian University of Technology\\
    \textsuperscript{\rm 2}School of Computer Science and Artificial Intelligence, Wuhan University of Technology\\
    \textsuperscript{\rm 3}School of Artificial Intelligence, Anhui University\\
    \textsuperscript{\rm 4}Anhui Provincial Key Laboratory of Multimodal Cognitive Computation, Anhui University\\

    \{924973292, 2981431354\}@mail.dlut.edu.cn, liuxuehu@whut.edu.cn, \\
    \{ly, zhpp, lhchuan\}@dlut.edu.cn, ahzheng214@foxmail.com
}

\usepackage{bibentry}

\begin{document}

\maketitle

\begin{abstract}
    Multi-modal object Re-IDentification (ReID) aims to retrieve specific objects by utilizing complementary image information from different modalities.
    Recently, large-scale pre-trained models like CLIP have demonstrated impressive performance in traditional single-modal object ReID tasks.
    However, they remain unexplored for multi-modal object ReID.
    Furthermore, current multi-modal aggregation methods have obvious limitations in dealing with long sequences from different modalities.
    To address above issues, we introduce a novel framework called MambaPro for multi-modal object ReID.
    To be specific, we first employ a Parallel Feed-Forward Adapter (PFA) for adapting CLIP to multi-modal object ReID.
    Then, we propose the Synergistic Residual Prompt (SRP) to guide the joint learning of multi-modal features.
    Finally, leveraging Mamba's superior scalability for long sequences, we introduce Mamba Aggregation (MA) to efficiently model interactions between different modalities.
    As a result, MambaPro could extract more robust features with lower complexity.
    Extensive experiments on three multi-modal object ReID benchmarks (i.e., RGBNT201, RGBNT100 and MSVR310) validate the effectiveness of our proposed methods.
    The source code is available at https://github.com/924973292/MambaPro.
  \end{abstract}

  \begin{figure}[t]
    \centering
    \includegraphics[width=0.45\textwidth]{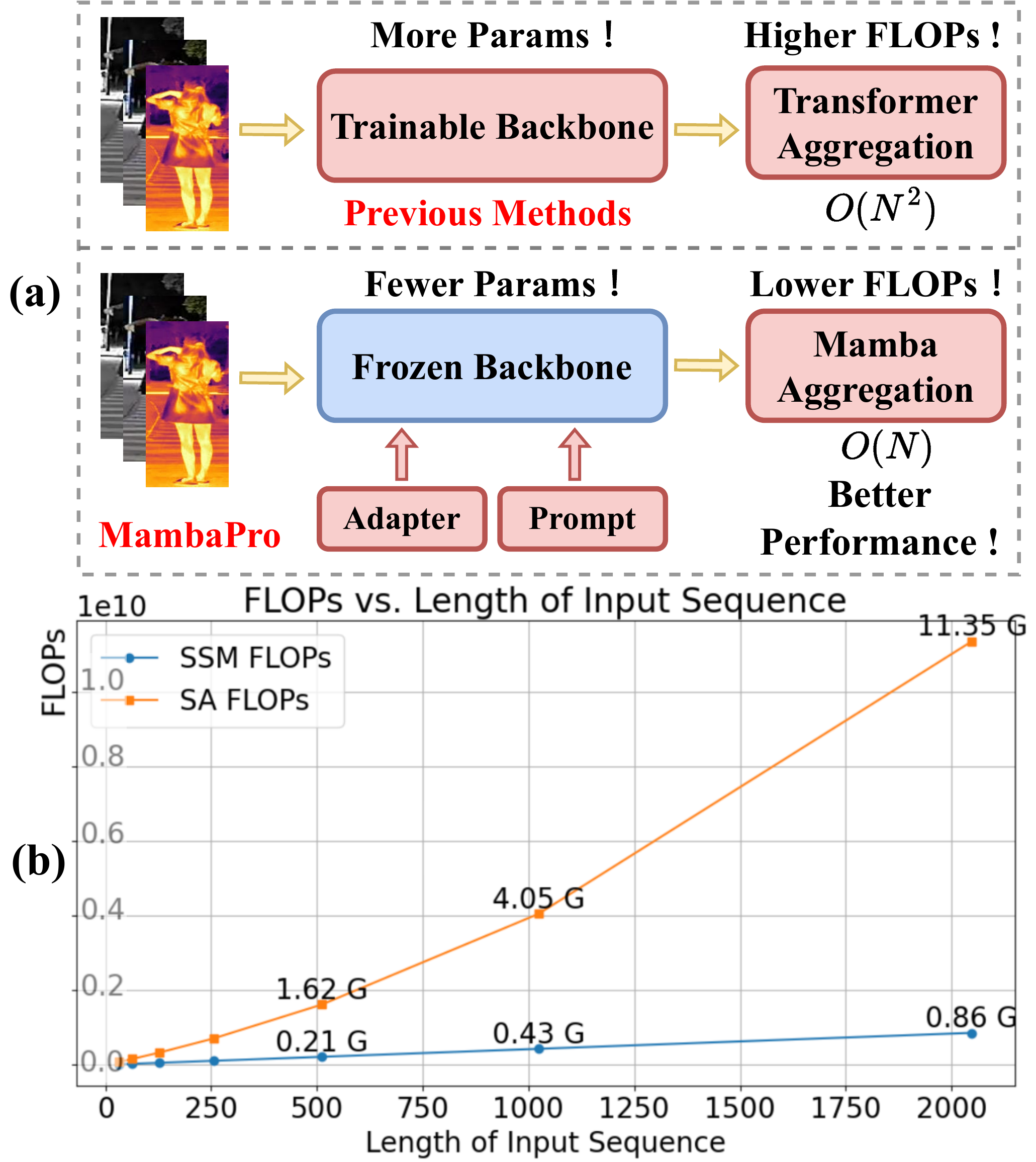}
    \vspace{-2mm}
    \caption{
      (a) Comparison between previous methods and MambaPro.
      (b) FLOPs comparison between SSM and SA.
    }
    \label{fig:Introduction}
    \vspace{-4mm}
  \end{figure}

  \section{Introduction}
  Object Re-IDentification (ReID) aims to re-identify specific objects across non-overlapping cameras.
  Due to its wide applications, object ReID has advanced significantly in recent years~\cite{liu2023deeply,wang2024other,liu2024video,liu2021watching,zhang2021hat,yu2024tf}.
  However, single-modal object ReID has many limitations in challenging scenarios~\cite{li2020multi,zheng2021robust}, such as lighting changes, shadows and low image resolutions.
  Under such extreme conditions, single-modal object ReID methods may extract misleading features~\cite{li2020multi}, resulting in the loss of discriminative information~\cite{wang2023top}.
  Fortunately, multi-modal object ReID has demonstrated a promising capability in addressing these challenges~\cite{wang2023top,shi2024multi,shi2024learning}.
  With complementary image information from different modalities, multi-modal object ReID methods can obtain more robust feature representations~\cite{zheng2023dynamic,wang2023top,crawford2023unicat,Yang_2023_ICCV,yang2024shallow}.
  Meanwhile, with the widespread application of Transformers~\cite{vaswani2017attention}, the framework for multi-modal object ReID also changes from CNN-based methods~\cite{wang2022interact,zheng2023dynamic} to Transformer-based methods~\cite{wang2023top,zhang2024magic}.
  Although these methods show promising performance, they still face challenges in computational complexity as shown in the top of Fig.~\ref{fig:Introduction} (a).

  Recently, large-scale pre-trained models like CLIP~\cite{radford2021learning} have demonstrated strong generalization capabilities in downstream tasks, such as single-modal object ReID~\cite{li2023clipreid}.
  However, full fine-tuning of CLIP leads to the high computational complexity and catastrophic forgetting~\cite{french1999catastrophic}.
  To address above issues, Parameter-Efficient Fine-Tuning (PEFT) methods like adapter and prompt tuning show competitive performance with fewer parameters and FLOPs.
  Meanwhile, current multi-modal aggregation methods face challenges in managing the high computational complexity that arises from interactions between different modalities.
  As shown in Fig.~\ref{fig:Introduction} (b), Self-Attention (SA)-based aggregation methods~\cite{wang2023top, yin2023graft, pan2023progressively} struggle with quadratic complexity when handling long sequences from multiple modalities.
  Fortunately, the Selective State Space Models (SSM) in Mamba~\cite{gu2023mamba} offer an efficient solution.
  They provide a superior scalability for long sequences by reducing complexity to linear levels~\cite{gu2021efficiently}.
  Motivated by the above observations, we propose MambaPro, a novel framework that combines CLIP-driven synergistic prompt tuning with Mamba aggregation.

  To be specific, our MambaPro consists of three key components: Parallel Feed-Forward Adapter (PFA), Synergistic Residual Prompt (SRP) and Mamba Aggregation (MA).
  Technically, we start by inserting the PFA into the frozen image encoder of CLIP to facilitate the transfer of pre-trained knowledge into the ReID task.
  To achieve this, the PFA is implemented as a parallel branch to the Feed-Forward Network (FFN).
  This parallel design not only maintains the original multi-modal feature representations but also supports flexible knowledge transfer across modalities.
  By combining features derived from the CLIP pre-trained model with refinements from the adapter, the framework can extract more robust representations for each modality.
  Then, we introduce the SRP to guide the joint learning of multi-modal features.
  To be specific, we propose a Synergistic Prompt (SP) to achieve the synergistic transfer with modality-specific prompts, facilitating effective exchanges of discriminative multi-modal information.
  Furthermore, we introduce a Residual Prompt (RP) to layer-wisely aggregate multi-modal information with residual refinements.
  As a result, our SRP can significantly enhance the information synergism.
  Finally, we propose the MA to efficiently model interactions between long sequences from different modalities.
  Specifically, the stacked MA blocks integrate complementary features from two aspects: intra-modality and inter-modality.
  With the cooperation of these two aspects, our MA can fully capture discriminative information with linear complexity.
  As a result, our framework can extract more robust multi-modal features with lower complexity.
  Experiments on three multi-modal object ReID benchmarks clearly demonstrate the effectiveness of our proposed methods.

  In summary, our contributions are as follows:
  \begin{itemize}
    \item
    We propose MambaPro, a novel framework for multi-modal object ReID.
    To our best knowledge, we are the first to introduce CLIP into multi-modal object ReID with Mamba aggregation and synergistic prompt tuning.
    \item
    We develop a Synergistic Residual Prompt (SRP) to guide the joint learning of multi-modal features, which effectively facilitate knowledge transfer and modality interactions with fewer parameters and FLOPs.
    \item
    We introduce a Mamba Aggregation (MA) to fully integrate the complementary information within and across different modalities with linear complexity.
    \item
    Extensive experiments are performed on three multi-modal object ReID benchmarks.
    The results fully validate the effectiveness of our proposed methods.
  \end{itemize}
  \section{Related Work}
  \subsection{Multi-Modal Object ReID}
  Traditional object ReID aims to extract features with single-modal input, such as RGB images or depth maps.
  In the past few years, single-modal object ReID has achieved remarkable progress.
  However, relying solely on single-modal input, these methods may extract limited information in complex scenarios.
  Meanwhile, the generalization capability of multi-modal data has sparked increasing attention in multi-modal object ReID~\cite{zheng2021robust,li2020multi,yin2023graft,he2023graph,wang2023top}.
  However, multi-modal input introduces additional challenges, such as distribution gaps among different modalities~\cite{liang2022mind} and modality laziness~\cite{crawford2023unicat}.
  To address above challenges, Zheng \emph{et al.}~\cite{zheng2021robust} propose to learn robust features through a progressive fusion for multi-modal person ReID.
  For multi-modal vehicle ReID, Li \emph{et al.}~\cite{li2020multi} propose to learn balanced multi-modal features.
  Based on Transformers~\cite{vaswani2017attention}, Pan \emph{et al.}~\cite{pan2023progressively} employ a feature hybrid mechanism to control modal-specific information.
  Recently, Zhang \emph{et al.}~\cite{zhang2024magic} propose to select diverse tokens with the cooperation of multi-modal information.
  Despite promising results, these methods still face challenges in generalization capability and model complexity.
  Especially for previous Transformer-based methods, full fine-tuning introduces high computational complexity.
  Thus, with the strong transfer capabilities of large-scale pre-trained models, we introduce the powerful CLIP into multi-modal object ReID with fewer trainable parameters and FLOPs.
  \begin{figure*}[tb]
    \centering
    \includegraphics[width=0.99\textwidth]{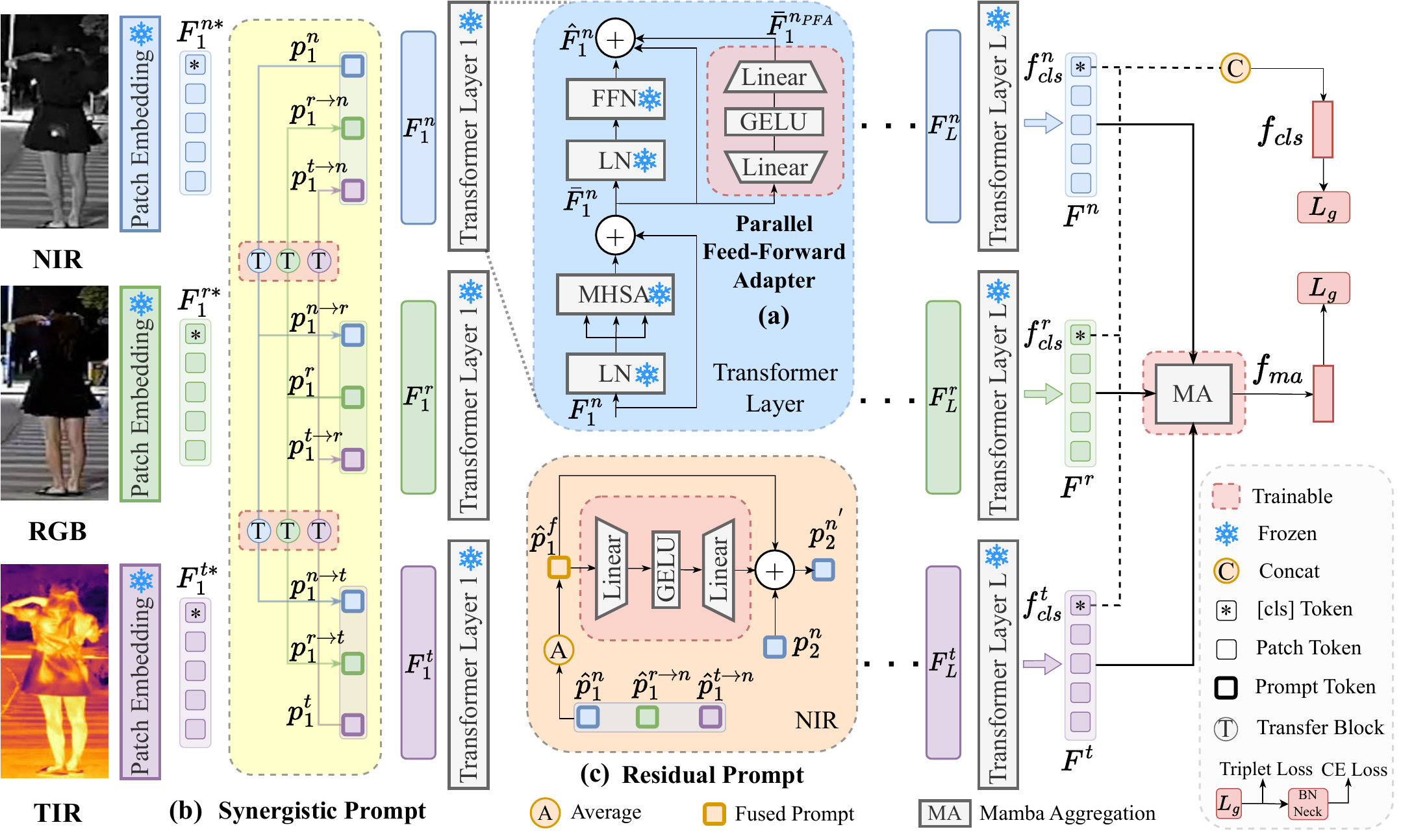}
    \caption{The overall framework of MambaPro.
    The Parallel Feed-Forward Adapter (PFA) is first introduced to transfer pre-trained knowledge into the ReID task.
    Then, the Synergistic Residual Prompt (SRP) is inserted to guide the progressive fusion of multi-modal features.
    Finally, the Mamba Aggregation (MA) is proposed to model interactions of long sequences from different modalities.
    With the proposed modules, our framework can obtain more robust features with low computational complexity.
    }
    \label{fig:Overall}
  \end{figure*}
  \subsection{Parameter-Efficient Fine-Tuning}
  With the development of foundational models~\cite{yang2024depth}, the transfer of pre-trained knowledge is attracting increasing attention~\cite{diao2025sherl,diao2024unipt}.
  Currently, mainstream PEFT methods can be divided into three categories: prompt tuning, adapter tuning and LoRA tuning.
  Prompt tuning~\cite{lester2021power} usually inserts layer-wise learnable tokens into the frozen backbone and guides the model to learn task-specific knowledge.
  Adapter tuning~\cite{houlsby2019parameter} introduces a plug-and-play module into the backbone, typically consisting of an MLP or attention module.
  LoRA tuning\cite{hu2021lora} uses a low-rank approximation as a side branch to reshape the backbone.
  With the above methods, PEFT demonstrates promising performance with limited resources.
  However, knowledge transfer for multi-modal fusion has been insufficiently explored, particularly in multi-modal object ReID.
  Therefore, we propose the PFA and SRP to leverage adapter tuning and prompt tuning for multi-modal object ReID.
  With the parallel adapters and synergistic transformations across multi-modal prompts, we can fully integrate the complementary information with lower complexity.
  \subsection{Visual State Space Models}
  Recently, Mamba~\cite{gu2023mamba} draws considerable attention~\cite{wan2024sigma,xu2024polyp} with the superior scalability of State Space Models (SSM)~\cite{gu2021combining}.
  By making the Structured State Space Sequence Model (S4)~\cite{gu2021efficiently} data-dependent with a selection mechanism, Mamba outperforms Transformers in handling long sequences with linear complexity.
  In computer vision, ViS4mer~\cite{islam2022long} and TranS4mer~\cite{islam2023efficient} demonstrate the effectiveness of SSM.
  To fully model the interactions between image patches, Vision Mamba~\cite{zhu2024vision} and VMamba~\cite{liu2024vmamba} employ different scanning strategies.
  However, most of these methods focus on single-modal tasks.
  Additionally, attention-based aggregation methods often face quadratic complexity.
  To address these issues, we design the intra-modality and inter-modality aggregation with Mamba, which together form the complete MA block.
  This approach integrates complementary information with linear complexity and complements the progressive fusion from PFA and SRP, enhancing feature robustness for ReID.
  \section{Method}
  As shown in Fig.~\ref{fig:Overall}, our framework comprises three main components: Parallel Feed-Forward Adapter (PFA), Synergistic Residual Prompt (SRP) and Mamba Aggregation (MA).
  With the frozen image encoder of CLIP as the shared backbone, we extract discriminative features from different modalities, i.e., RGB, Near Infrared (NIR) and Thermal Infrared (TIR).
  Details of the each component are as follows.
  \subsection{Parallel Feed-Forward Adapter}
  To transfer CLIP's knowledge to ReID tasks, we introduce the Parallel Feed-Forward Adapter (PFA).
  Previous methods~\cite{diao2024unipt} typically insert adapters sequentially into networks.
  However, this can disrupt the original information flow, leading to suboptimal feature transformations.
  Especially in multi-modal settings, it is crucial to maintain the integrity of input features.
  To address these issues, our PFA is designed as a parallel branch to the Feed-Forward Network (FFN) rather than as a sequential operation.
  Taking the NIR modality as an example, as shown in Fig.~\ref{fig:Overall} (a), the tokenized features \(F_{1}^{n}\) are first processed through the initial Transformer layer with the following equation:
  \begin{equation}
    \bar{F}^{n}_{1} = \mathcal{M}(\mathcal{L}({F}^{n}_{1})) + {F}^{n}_{1},
  \end{equation}
  where \(\mathcal{L}\) represents the Layer Normalization (LN)~\cite{ba2016layer} and \(\mathcal{M}\) denotes the multi-head self-attention~\cite{dosovitskiy2020image}.
  The output \(\bar{F}^{n}_{1}\) is then fed into the PFA, undergoing the following transformation:
  \begin{equation}
  \bar{F}^{n_{PFA}}_{1} = \xi(\delta(\xi(\bar{F}^{n}_{1}))).
  \end{equation}
  Here, \(\xi\) is a linear layer and \(\delta\) is the GELU activation function~\cite{hendrycks2016gaussian}.
  Finally, we integrate the output \(\bar{F}^{n_{PFA}}_{1}\) with the FFN, resulting in the feature \(\hat{F}^{n}_{1}\):
  \begin{equation}
  \hat{F}^{n}_{1} = \mathcal{F}(\mathcal{L}(\bar{F}^{n}_{1})) + \bar{F}^{n_{PFA}}_{1} + \bar{F}^{n}_{1},
  \end{equation}
  where \(\mathcal{F}\) represents the FFN.
  This parallel structure preserves the original feature representations while enabling flexible and efficient knowledge transfer across different modalities.
  Additionally, PFA employs an ascending-then-descending pattern in linear layers.
  This design enhances its ability to adapt to complex multi-modal feature distributions, supporting more stable knowledge transfer.
  \subsection{Synergistic Residual Prompt}
  To promote the joint learning of multi-modal features, we introduce the Synergistic Residual Prompt (SRP).
  Different from previous single modal prompt-based methods, our SRP is designed to integrate the modality-specific knowledge with synergistic transformations.
  Besides, current multi-modal prompt-based methods~\cite{li2023clip,khattak2023maple} often discard prompts from earlier layers, potentially resulting in the loss of discriminative information.
  To address these issues, our SRP introduces two key components: Synergistic Prompt (SP) and Residual Prompt (RP).
  The SP transfers modality-specific knowledge across different modalities, while the RP aggregates multi-modal information across layers with residual refinements.
  \\
  \textbf{Synergistic Prompt.}
  Without loss of generality, we take the NIR modality as an example.
  In Fig.~\ref{fig:Overall} (b), the NIR image is first tokenized by the patch embedding~\cite{vaswani2017attention}.
  The patch tokens $f_\mathrm{pa}^{n} \in \mathbb{R}^{D \times N_{pa}}$ are concatenated with the class token $f_\mathrm{cls}^{n} \in \mathbb{R}^{D}$ to form $F_{1}^{n*} \in \mathbb{R}^{D \times (1+N_{pa})}$:
  \begin{equation}
    F_{1}^{n*} = \left[f_\mathrm{cls}^{n},f_\mathrm{pa}^{n}\right],
  \end{equation}
  where $[\cdot]$ is the concatenation operation.
  For the prompt tokens in NIR modality, they are composed of three parts: $p_\mathrm{1}^{n}, p_\mathrm{1}^{r \to n}, p_\mathrm{1}^{t \to n} \in \mathbb{R}^{D \times N_{pr}}$.
  Here, $p_\mathrm{1}^{n}$ is the randomly initialized NIR-specific prompt, while $p_\mathrm{1}^{r \to n}$ and $p_\mathrm{1}^{t \to n}$ are transferred from RGB and TIR modalities, respectively.
  Finally, all of them will be concatenated to form the input $F_{1}^{n} \in \mathbb{R}^{D \times (1+N_{pa}+3N_{pr})}$ of the first layer as follows:
  \begin{equation}
    F_{1}^{n} = \left[F_{1}^{n*},p_\mathrm{1}^{n}, p_\mathrm{1}^{r \to n}, p_\mathrm{1}^{t \to n}\right].
  \end{equation}
  Here, $D$ is the embedding dimension and $N_{pa}$ is the number of patches.
  $N_{pr}$ is the number of prompts in each modality.
  \begin{equation}
    p_\mathrm{1}^{x \to n} = \mathcal{T}(p_{1}^{x})=\xi(\delta(\xi(p_\mathrm{1}^{x}))),
  \end{equation}
  where \( x \) can be \( r \) or \( t \).
  \(\mathcal{T}\) is the transfer block composed of linear layers and a GELU activation function.
  Then, the output $\hat{F}_{1}^{n}$ of the first Transformer layer $\Phi_{1}$ is calculated with the input $F_{1}^{n}$ with the following equation:
  \begin{equation}
    \hat{F}_{1}^{n} = \mathrm{\Phi_{1}}(F_{1}^{n}).
  \end{equation}
  Other modalities undergo similar operations.
  As a result, we can leverage SP to facilitate the transfer of discriminative modality-specific information across different modalities.
  \\
  \textbf{Residual Prompt.}
  As shown in Fig.~\ref{fig:Overall} (c), we first extract prompt tokens $\hat{p}_{1}^{n}, \hat{p}_{1}^{r \to n}, \hat{p}_{1}^{t \to n}$ from $\hat{F}_{1}^{n}$, with NIR modality as an example.
  Then, we average these prompts to form the fused prompts $\hat{p}_{1}^{f} \in \mathbb{R}^{D \times N_{pr}}$.
  Afterwords, we add the fused prompts $\hat{p}_{1}^{f}$ with linear layers to the new NIR-specific prompt $p_{2}^{n}$ to form the final prompts for the second layer:
  \begin{equation}
    p_{2}^{n^{'}} =  p_{2}^{n} + \xi(\delta(\xi(\hat{p}_{1}^{f}))),\hat{p}_{1}^{f} = \frac{\hat{p}_{1}^{n}+\hat{p}_{1}^{r \to n}+\hat{p}_{1}^{t \to n}}{3}.
  \end{equation}
  Finally, the input $F_{2}^{n}$ of the second layer $\Phi_{2}$ is calculated as:
  \begin{equation}
    F_{2}^{n} = \left[F_{2}^{n*},p_{2}^{n^{'}}, \mathcal{T}(p_{2}^{r}),  \mathcal{T}(p_{2}^{t})\right].
  \end{equation}
  %
  %
  Other modalities and layers are processed in the same way:
  \begin{equation}
    \hat{F}_{l}^{n} = \mathrm{\Phi_{\mathnormal{l}}}(F_{l}^{n}),
    F_{l}^{n} = \left[F_{l}^{n*},p_{l}^{n^{'}}, \mathcal{T}(p_{l}^{r}), \mathcal{T}(p_{l}^{t})\right],
  \end{equation}
  \begin{equation}
    \hat{F}_{l}^{r} = \mathrm{\Phi_{\mathnormal{l}}}(F_{l}^{r}),
    F_{l}^{r} = \left[F_{l}^{r*},\mathcal{T}(p_{l}^{n}), p_{l}^{r^{'}}, \mathcal{T}(p_{l}^{t})\right],
  \end{equation}
  \begin{equation}
    \hat{F}_{l}^{t} = \mathrm{\Phi_{\mathnormal{l}}}(F_{l}^{t}),
    F_{l}^{t} = \left[F_{l}^{t*},\mathcal{T}(p_{l}^{n}), \mathcal{T}(p_{l}^{r}), p_{l}^{t^{'}}\right],
  \end{equation}
  where $l \in \{2,3,...,L\}$.
  With the  synergistic transfer of modality-specific knowledge across different modalities and the residual refinements of multi-modal information, our SRP fully enhances the information synergism, promoting the feature discrimination.
  After the feature extraction from the backbone, we obtain $F^n, F^r ,F^t \in \mathbb{R}^{D \times (1+N_{pa})}$ for each modality.
  Then, we extract class tokens $f^{n}_{cls},f^{r}_{cls},f^{t}_{cls}$ and concatenate them to form $f_{cls} \in \mathbb{R}^{3D}$ for supervision.
  \begin{figure}[tb]
    \centering
    \includegraphics[width=0.47\textwidth]{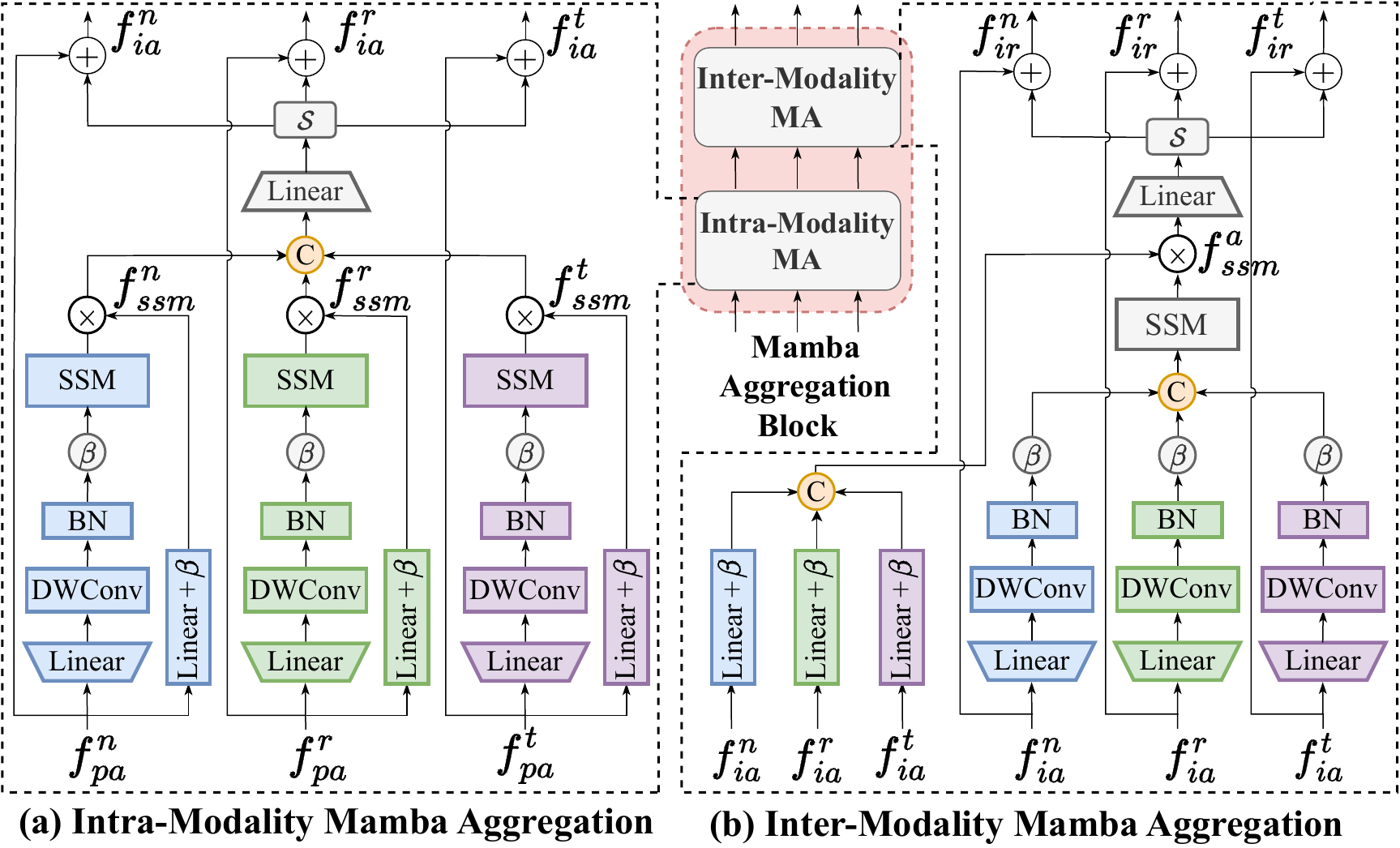}
    \vspace{-4mm}
    \caption{Details of our proposed Mamba Aggregation.}
    \label{fig:MA_block}
    \vspace{-0mm}
  \end{figure}
  \subsection{Mamba Aggregation}
  To efficiently model interactions between tokenized sequences from different modalities, we introduce Mamba Aggregation (MA) as a fusion strategy.
  As depicted in Fig.~\ref{fig:MA_block}, MA is essentially composed of two parts: intra-modality MA and inter-modality MA.
  The intra-modality MA is designed to capture the discriminative information within each modality.
  While the inter-modality MA is introduced to fully integrate the complementary features across different modalities.
  In the intra-modality MA, we first extract patch tokens $f_{pa}^{n}, f_{pa}^{r}, f_{pa}^{t} \in \mathbb{R}^{D \times N_{pa}}$ from $F^n$, $F^r$ and $F^t$, respectively.
  In the left part of Fig. \ref{fig:MA_block}, each of them will undergo the Mamba block composed of Depth-Wise Convolution (DWConv)~\cite{chollet2017xception} $\mathcal{D}$, Batch Normalization (BN)~\cite{ioffe2015batch} $\mathcal{B}$, SSM $\Omega$ and SiLU activation function~\cite{elfwing2018sigmoid} $\beta$ with the following transformations:
  \begin{equation}
    \mathrm{\Theta}(\mathrm{\mathcal{X}}) = \beta(\mathcal{B}(\mathcal{D}(\xi (\mathcal{X})))),\mathrm{\Psi}(\mathcal{X}) = \mathrm{\beta}(\xi (\mathcal{X})),
  \end{equation}
  \begin{equation}
    f^M_{ssm} = \Omega (\mathrm{\Theta}(f_{pa}^{M})) \times \mathrm{\Psi}(f_{pa}^{M}), M \in \{n,r,t\}.
  \end{equation}
  Here, $M$ denotes the modality.
  Then, features from different Mamba blocks are concatenated along the token dimension.
  After passing through a linear layer, the aggregated features are split by modality for the intra-modality MA:
  \begin{equation}
    \scalebox{0.87}{$
    f_{ia}^{n}, f_{ia}^{r}, f_{ia}^{t} = \mathcal{S}\left(\xi\left(\left[f^n_{ssm}, f^r_{ssm}, f^t_{ssm}\right]\right) + \left[f_{pa}^{n}, f_{pa}^{r}, f_{pa}^{t}\right]\right),
    $}
  \end{equation}
  where $f_{ia}^{n}, f_{ia}^{r}, f_{ia}^{t} \in \mathbb{R}^{D \times N_{pa}}$.
  Here, $\mathcal{S}$ denotes the feature splitting along the token dimension.
  Thus, through a separable SSM for each modality, the intra-modality MA effectively captures interactions within each modality.

  Furthermore, as shown in the right part of Fig. \ref{fig:MA_block}, $f_{ia}^{n}$, $f_{ia}^{r}$ and $f_{ia}^{t}$ will be sent to the inter-modality MA as follows:
  \begin{equation}
    \scalebox{0.81}{$f_{ssm}^{a} = \Omega([\mathrm{\Theta}(f_{ia}^{n}), \mathrm{\Theta}(f_{ia}^{r}), \mathrm{\Theta}(f_{ia}^{t})]) \times [\mathrm{\Psi}(f_{ia}^{n}), \mathrm{\Psi}(f_{ia}^{r}), \mathrm{\Psi}(f_{ia}^{t})].
    $}
  \end{equation}
  Here, $f_{ssm}^{a} \in \mathbb{R}^{D \times 3N_{pa}}$ represents the aggregated features obtained by modeling all modality tokens as the long sequences.
  Finally, we split the aggregated features into different modalities for the next MA block:
  \begin{equation}
    f_{ir}^{n}, f_{ir}^{r}, f_{ir}^{t} = \mathcal{S}(\xi(f_{ssm}^{a}) + [f_{ia}^{n}, f_{ia}^{r}, f_{ia}^{t}]).
  \end{equation}
  Through the cooperation of the intra-modality MA and inter-modality MA, MA fully models the interactions of patch tokens within and across modalities.
  After the stacked MA blocks, we concatenate the class tokens and the average of patch tokens of each modality.
  Then, we use a LN to stabilize the feature learning.
  With a linear reduction, we obtain the final features $f_{final}^{M}, M \in \{n,r,t\}$ for each modality:
  \begin{equation}
    f_{final}^{M} =  \xi(\mathcal{L}(\left[f_{cls}^{M}, \mathcal{A}(f_{ir}^{M})\right])).
  \end{equation}
  Here, $\mathcal{A}$ represents the average operation.
  Finally, we concatenate the features from different modalities to form $f_{ma} \in \mathbb{R}^{3D}$ for the loss supervision:
  \begin{equation}
    f_{ma} =  [f_{final}^{n}, f_{final}^{r}, f_{final}^{t}].
  \end{equation}
  With the above aggregation strategy, we can fully model the interactions across modalities with linear complexity.
  \subsection{Objective Functions}
  As illustrated in Fig.~\ref{fig:Overall}, our objective function comprises two parts: losses for the image encoder of CLIP and the MA.
  For the backbone and MA, they are both supervised by the label smoothing cross-entropy loss~\cite{szegedy2016rethinking} and triplet loss~\cite{hermans2017defense}:
  \begin{equation}
    \mathcal{L}_{g} = \mathrm{\lambda_1}\mathcal{L}_{ce} + \mathrm{\lambda_2}\mathcal{L}_{tri}.
  \end{equation}
  Finally, the total loss for our framework can be defined as:
  \begin{equation}
  \begin{split}
  \mathcal{L}_{total} = \mathcal{L}_{g}^{CLIP} + \mathcal{L}_{g}^{MA}.
  \end{split}
  \end{equation}
  \section{Experiments}
  \subsection{Experimental Setup}
  \textbf{Datasets and Evaluation Protocols.}
  To fully evaluate the performance of our method, we conduct experiments on three multi-modal object ReID benchmarks.
  Specifically, RGBNT201~\cite{zheng2021robust} is a multi-modal person ReID dataset comprising RGB, NIR and TIR images.
  RGBNT100~\cite{li2020multi} is a large-scale multi-modal vehicle ReID dataset with diverse visual challenges, such as abnormal lighting, glaring and occlusion.
  MSVR310~\cite{zheng2022multi} is a small-scale multi-modal vehicle ReID dataset with more complex visual challenges.
  For evaluation metrics, we utilize the mean Average Precision (mAP) and Cumulative Matching Characteristics (CMC) at Rank-K ($K=1,5,10$).
  Meanwhile, we report the GPU memory, trainable parameters and FLOPs for complexity analysis.
  \\
  \textbf{Implementation Details.}
  Our model is implemented by using the PyTorch toolbox with one NVIDIA A100 GPU.
  We employ the pre-trained image encoder of CLIP~\cite{radford2021learning} as the backbone.
  For the input resolution, images are resized to 256$\times$128 for RGBNT201 and 128$\times$256 for RGBNT100/MSVR310.
  For data augmentation, we employ random horizontal flipping, cropping and erasing~\cite{zhong2020random}.
  For small-scale datasets (i.e., RGBNT201 and MSVR310), the mini-batch size is set to 64, with 4 images sampled for each identity and 16 identities sampled in a batch.
  For the large-scale dataset, i.e., RGBNT100, the mini-batch size is set to 128, with 16 images sampled for each identity.
  We set $\mathrm{\lambda_1}$ to 0.25 and $\mathrm{\lambda_2}$ to 1.0.
  We use the Adam optimizer to fine-tune the model with a learning rate of 3.5$\mathrm{e}^{-4}$.
  The warmup strategy with a cosine decay is used for learning rate scheduling.
  We set the total number of training epochs to 60 for RGBNT201/MVSR310 and 30 for RGBNT100, respectively.
  \begin{table}[tb]
    \centering
    \renewcommand\arraystretch{1.15}
    \setlength\tabcolsep{4pt}
    \resizebox{0.47\textwidth}{!}
  {
    \begin{tabular}{cccccc}
      \noalign{\hrule height 1pt}
      \multicolumn{2}{c}{\multirow{2}{*}{Methods}}     &\multicolumn{4}{c}{RGBNT201}\\ \cline{3-6}
      & & mAP & R-1 & R-5 & R-10 \\
      \hline
      \multirow{4}{*}{\rotatebox{90}{Single}}
      &MUDeep~\cite{qian2017multi} & 23.8 & 19.7 & 33.1 & 44.3 \\
      &PCB~\cite{sun2018beyond}  & 32.8 & 28.1 & 37.4 & 46.9 \\
      &OSNet~\cite{zhou2019omni} & 25.4 & 22.3 & 35.1 & 44.7 \\
      &CAL~\cite{rao2021counterfactual}  & 27.6 & 24.3 & 36.5 & 45.7 \\
      \hline
      \multirow{11}{*}{\rotatebox{90}{Multi}}
      & HAMNet~\cite{li2020multi}   & 27.7         & 26.3            & 41.5            & 51.7             \\
      & PFNet~\cite{zheng2021robust}    & 38.5         & 38.9            & 52.0              & 58.4             \\
      & IEEE~\cite{wang2022interact}     & 49.5         & 48.4            & 59.1           &65.6             \\
      & DENet~\cite{zheng2023dynamic}    & 42.4         & 42.2            & 55.3            & 64.5            \\
      & LRMM~\cite{wu2025lrmm} & 52.3 & 53.4 & 64.6 & 73.2\\
      & UniCat$^*$~\cite{crawford2023unicat}    & 57.0         & 55.7            & -            & -            \\
      & HTT$^*$~\cite{wang2024heterogeneous} &71.1 &73.4 &83.1 &87.3\\
      & TOP-ReID$^*$~\cite{wang2023top}  &\underline{72.3} &\underline{76.6} &\underline{84.7} &\underline{89.4}\\
      & EDITOR$^*$~\cite{zhang2024magic} & 66.5       & 68.3           & 81.1        & 88.2             \\
      & RSCNet$^*$~\cite{yu2024representation} & 68.2 & 72.5 & - & - \\
      \rowcolor[gray]{0.92}
      & MambaPro$\ddagger$ & \textbf{78.9} & \textbf{83.4} & \textbf{89.8} & \textbf{91.9} \\
      \noalign{\hrule height 1pt}
      \end{tabular}
      }
    \vspace{-1mm}
    \caption{Performance comparison on RGBNT201.
    The best results are highlighted in bold, with the second-best underlined.
    Symbols: $\ddagger$ indicates CLIP-based methods, $*$ marks ViT-based methods and others are CNN-based methods.}
    \label{tab:multi-spectral person ReID}
    \vspace{-2mm}
  \end{table}
  \subsection{Performance Comparison}
  \textbf{Multi-modal Person ReID.}
  As shown in Tab. \ref{tab:multi-spectral person ReID}, we compare our proposed MambaPro with state-of-the-art methods on RGBNT201.
  %
  %
  Within single-modal methods, PCB exhibits a remarkable mAP of 32.8\%.
  For multi-modal methods, CNN-based methods like IEEE~\cite{wang2022interact} and LRMM~\cite{wu2025lrmm} achieve competitive performance with mAPs of 49.5\% and 52.3\%, respectively.
  However, Transformer-based methods outperform CNN-based methods.
  Especially, TOP-ReID~\cite{wang2023top} achieves a mAP of 72.3\%.
  Moreover, our MambaPro achieves a 78.9\% mAP and 83.4\% Rank-1, surpassing TOP-ReID by 6.6\% and 6.8\%, respectively.
  These performance improvements verify the effectiveness of our proposed method.
  \\
  \textbf{Multi-modal Vehicle ReID.}
  As depicted in Tab. \ref{tab:multi-spectral vehicle ReID}, single-modal methods generally under-perform multi-modal methods.
  In single-modal methods, CNN-based methods like BoT~\cite{luo2019bag} achieve better results than some Transformer-based methods.
  For multi-modal methods, LRMM~\cite{wu2025lrmm} achieves a mAP of 78.6\% on RGBNT100, showing the effectiveness of its low rank fusion strategy.
  Meanwhile, Transformer-based methods exhibit their superiority on integrating multi-modal information.
  Specifically, EDITOR~\cite{zhang2024magic} achieves a mAP of 82.1\%, surpassing LRMM by 3.5\% on RGBNT100.
  Our MambaPro achieves a mAP of 83.9\% on RGBNT100, surpassing EDITOR by 1.8\%.
  Especially on the small-scale dataset MSVR310, our MambaPro achieves a mAP of 47.0\%, surpassing EDITOR by 8.0\%.
  These results clearly verify the effectiveness of our proposed methods in multi-modal object ReID tasks.
  \begin{table}[tb]
  \centering
  \renewcommand\arraystretch{1.15}
  \setlength\tabcolsep{5pt}
  \resizebox{0.436\textwidth}{!}
  {
  \begin{tabular}{cccccc}
    \noalign{\hrule height 1pt}
    \multicolumn{2}{c}{\multirow{2}{*}{Methods}} &  \multicolumn{2}{c}{RGBNT100} & \multicolumn{2}{c}{MSVR310} \\
    \cline{3-6}
    & & mAP & R-1 & mAP & R-1 \\
    \hline
    \multirow{4}{*}{\rotatebox{90}{single}}
    &PCB~\cite{sun2018beyond}& 57.2 & 83.5 & 23.2 & 42.9 \\
    &BoT~\cite{luo2019bag} & 78.0 & 95.1 & 23.5 & 38.4 \\
    &OSNet~\cite{zhou2019omni}& 75.0 & 95.6 & 28.7 & 44.8 \\
    &TransReID$^*$~\cite{he2021transreid}& 75.6 & 92.9 & 18.4 & 29.6 \\
    \hline
    \multirow{14}{*}{\rotatebox{90}{Multi}}
    &HAMNet~\cite{li2020multi}  & 74.5 & 93.3 & 27.1 & 42.3 \\
    &PFNet~\cite{zheng2021robust}& 68.1 & 94.1 & 23.5 & 37.4 \\
    &GAFNet~\cite{guo2022generative} & 74.4 & 93.4 & - & - \\
    &GPFNet~\cite{he2023graph} & 75.0 & 94.5 & - & - \\
    &CCNet~\cite{zheng2022multi} & 77.2 & 96.3 & 36.4 & \underline{55.2} \\
    & LRMM~\cite{wu2025lrmm} & 78.6 & \textbf{96.7} & 36.7 &49.7\\
    &GraFT$^*$~\cite{yin2023graft} &76.6 &94.3 &- &-\\
    &PHT$^*$~\cite{pan2023progressively}& 79.9 & 92.7 & - & - \\
    &UniCat$^*$~\cite{crawford2023unicat}    & 79.4         & 96.2  & -            & -            \\
    & HTT$^*$~\cite{wang2024heterogeneous} &75.7&92.6&- &-\\
    & TOP-ReID$^*$~\cite{wang2023top} &81.2 & 96.4 & 35.9 & 44.6 \\
    & EDITOR$^*$~\cite{zhang2024magic} & 82.1 & 96.4 &39.0 & 49.3\\
    & RSCNet$^*$~\cite{yu2024representation} &\underline{82.3} &\underline{96.6} &\underline{39.5} &49.6\\
    \rowcolor[gray]{0.92}
    & MambaPro$\ddagger$ & \textbf{83.9} & 94.7 & \textbf{47.0} & \textbf{56.5} \\
    \noalign{\hrule height 1pt}
    \end{tabular}}
  \vspace{-2mm}
  \caption{Performance on RGBNT100 and MSVR310.}
  \label{tab:multi-spectral vehicle ReID}
  \vspace{-4mm}
  \end{table}
  \subsection{Ablation Study}
  We conduct ablation studies on RGBNT201 to validate the effectiveness of different components.
  The locked image encoder of CLIP is denoted as CLIP(L) with frozen parameters.
  Meanwhile, we refer to the fully fine-tuned CLIP as CLIP(F).
  When MA is not applied, we use $f_{cls}$ for retrieval.
  Otherwise, we use $f_{ma}$ for retrieval.
  \\
  \textbf{Effect of Key Components.}
  Tab. \ref{tab:ablation} shows the performance comparison with different components.
  Comparing Model A with Model B, we observe an impressive mAP improvement by fully fine-tuning CLIP, demonstrating the necessity of fine-tuning.
  With 32.9\% of the trainable parameters compared with Model B, Model C outperforms the fully fine-tuned CLIP by 3.8\% mAP, showing the effectiveness of PFA.
  Meanwhile, Model D also surpasses Model B with minimal additional parameters and FLOPs.
  Furthermore, through the combination of SRP and PFA, Model E achieves higher performance.
  Besides, with the efficient modeling of multi-modal long sequences, Model F achieves the best mAP of 78.9\%.
  Model G, with the full fine-tuning of CLIP, achieves a comparable mAP of 78.7\%.
  Comparing the last two rows, our method achieves robust performance no matter whether the backbone is fully fine-tuned or not.
  With our efficient learning framework, MambaPro transfers the knowledge from CLIP to multi-modal object ReID with superior performance and low complexity.
  \begin{table}[t]
    \centering
    \vspace{-0mm}
    \renewcommand\arraystretch{1.15}
    \setlength\tabcolsep{2pt}
    \resizebox{0.47\textwidth}{!}
    {
    \begin{tabular}{c|ccccc|cccc|cc}
      \noalign{\hrule height 1pt}
        & \multicolumn{5}{c|}{Methods} & \multicolumn{4}{c|}{Performance} & \multicolumn{2}{c}{Complexity} \\
          \cline{2-12}
          &CLIP(L)&CLIP(F)&PFA & SRP &MA & mAP & R-1 & R-5 & R-10 & Mem(GB) & Params(M) \\
        \hline
        A&\ding{51}&\ding{53}&\ding{53}&\ding{53}&\ding{53}& 3.3 & 2.8 & 7.4 & 11.1 &12.79&0 \\
        B&\ding{53}&\ding{51}&\ding{53}&\ding{53}&\ding{53}& 69.4 & 71.5 & 81.8 & 87.4&16.85&86.14 \\
        C&\ding{51}&\ding{53}&\ding{51}&\ding{53}&\ding{53}& 73.2 & 76.4 & 85.2 & 88.2&15.96&28.34\\
        D&\ding{51}&\ding{53}&\ding{53}&\ding{51}&\ding{53}& 71.3 & 74.5 & 83.4 & 87.9&13.82&28.48 \\
        E&\ding{51}&\ding{53}&\ding{51}&\ding{51}&\ding{53}& 74.4 & 77.6 & 87.4 & 90.3&17.79&56.82 \\
        \rowcolor[gray]{0.92}
        \textbf{F}&\ding{51}&\ding{53}&\ding{51}&\ding{51}&\ding{51}& \textbf{78.9} & \underline{83.4} & \textbf{89.8} & \underline{91.9}&20.89&74.20 \\
        G&\ding{53}&\ding{51}&\ding{51}&\ding{51}&\ding{51}& \underline{78.7} & \textbf{83.5} & \underline{89.6} & \textbf{92.9}&25.22&160.34 \\
        \noalign{\hrule height 1pt}
    \end{tabular}
    }
    \vspace{-2mm}
    \caption{Performance comparison with key components.}
    \vspace{-4mm}
    \label{tab:ablation}
  \end{table}
  \begin{figure}[t]
    \centering
    \includegraphics[width=0.34\textwidth]{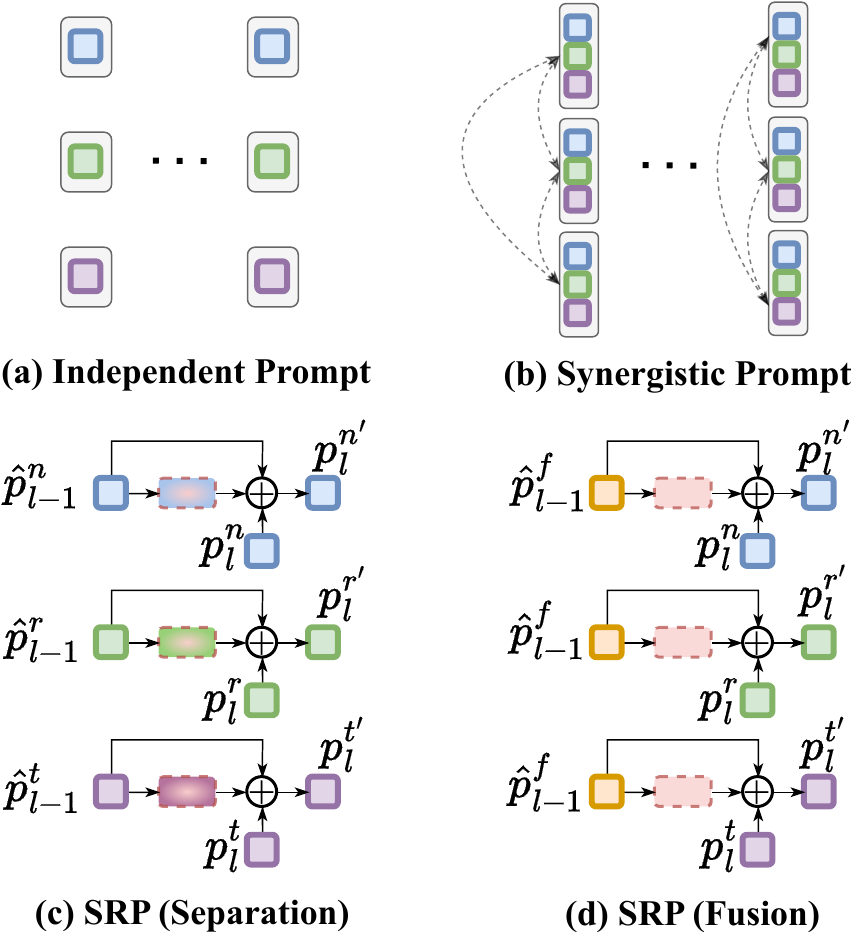}
    \vspace{-2mm}
    \caption{Details of different prompt mechanisms.}
    \label{fig:prompt}
    \vspace{-6mm}
  \end{figure}
  \begin{table}[t]
    \centering
    \vspace{-0mm}
    \renewcommand\arraystretch{1.15}
    \setlength\tabcolsep{1.5pt}
    \resizebox{0.48\textwidth}{!}
    {
    \begin{tabular}{c|cccc|cc}
      \noalign{\hrule height 1pt}
        {\multirow{2}{*}{Methods}} & \multicolumn{4}{c|}{Performance} & \multicolumn{2}{c}{Prompt Complexity} \\\cline{2-7}
        & mAP & R-1 & R-5 & R-10 & FLOPs(G) & Params(M) \\
        \hline
        CLIP(L) + IP & 64.7 & 65.9 & 77.3 & 82.9 &0.09&0.01\\
        CLIP(L) + SP & 68.9 & 70.7 & 80.9 & 85.2 &0.28&1.78\\
        CLIP(L) + SRP(Separation) & 70.1 & 72.3 & 82.1 & 86.7 &0.29&3.55\\
        \rowcolor[gray]{0.92}
        \textbf{CLIP(L) + SRP(Fusion)} & \textbf{71.3} & \textbf{74.5} & \textbf{83.4} & \textbf{87.9} &0.29&2.37\\
        \noalign{\hrule height 1pt}
    \end{tabular}}
    \vspace{-3mm}
    \caption{Comparison with different prompt mechanisms.}
    \label{tab:prompt_comparison}
    \vspace{-2mm}
  \end{table}
  \begin{figure*}[t]
    \vspace{-0mm}
    \centering
    \includegraphics[width=0.9\textwidth]{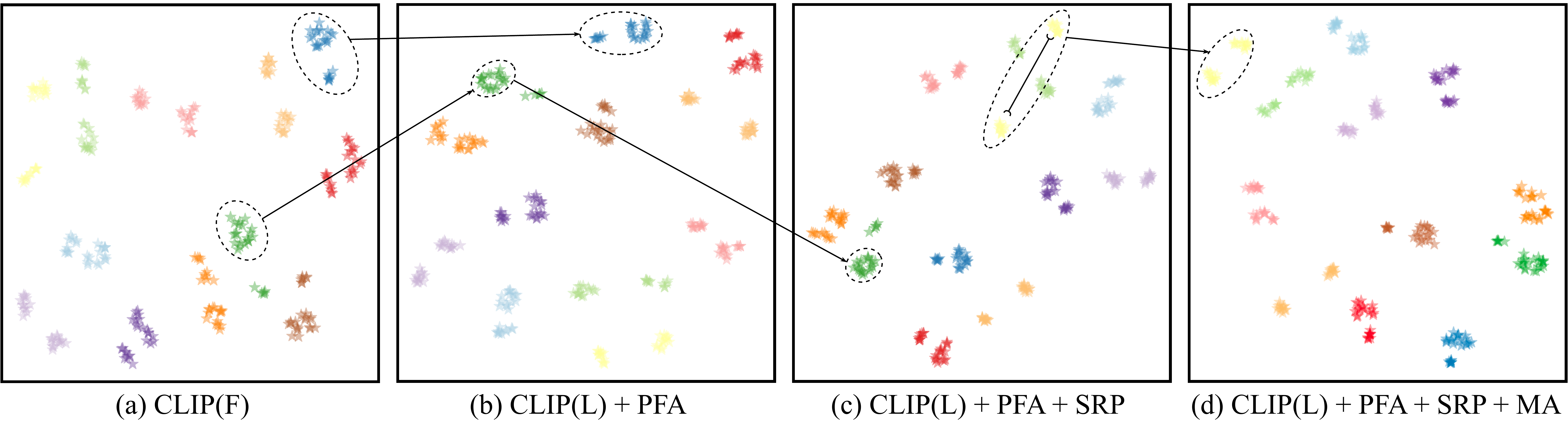}
    \vspace{-3mm}
    \caption{Feature distributions with t-SNE~\cite{van2008visualizing}.
    Different colors represent different IDs.
    }
    \label{fig:tsne}
    \vspace{-4mm}
  \end{figure*}
  \\
  \textbf{Effect of Different Prompt Mechanisms.}
  Fig. \ref{fig:prompt} illustrates the details of different prompt mechanisms.
  Tab. \ref{tab:prompt_comparison} shows the corresponding performances.
  The first row represents that each modality has its own prompts without synergistic transfer.
  SP achieves better performance than Independent Prompt (IP), demonstrating the effectiveness of synergistic transformation between modalities.
  Both SRP (Separation) and SRP (Fusion) outperform SP, underscoring the effectiveness of residual connections in multi-modal prompts.
  The difference between SRP (Separation) and SRP (Fusion) lies in the $\hat{p}_{l}^{f}$ comes from the prompts of separate modalities or the fusion of them.
  SRP (Fusion) performs best by averaging tokens with the residual connections.
  \\
  \textbf{Effect of Different Adapters.}
  Tab. \ref{tab:adapter_comparison} shows the performance with different adapters.
  Due to the weaker learning capability without non-linear transformations, LoRA generally achieves inferior performance compared with the Bottleneck Adapter (BNA).
  Meanwhile, with acceptable complexity, our PFA outperforms BNA by 4.7\% mAP, verifying the effectiveness in multi-modal object ReID.
  \\
  \textbf{Effect of Different Aggregation Methods.}
  In Tab. \ref{tab:Main_fusion_comparison}, we compare the performance with different aggregation methods.
  The Sum and Concat are simple aggregation methods that yield inferior results.
  The Transformer achieves better performance, demonstrating the effectiveness of attention-based aggregation.
  Our Intra-MA and Inter-MA outperform Transformers, showcasing the effectiveness of modeling multi-modal sequences with Mamba.
  Finally, our MA achieves the best performance, demonstrating the effectiveness of our proposed method.
  \begin{table}[t]
    \centering
    \vspace{-0mm}
    \renewcommand\arraystretch{1.15}
    \setlength\tabcolsep{1.5pt}
    \resizebox{0.47\textwidth}{!}
    {
    \begin{tabular}{c|cccc|cc}
      \noalign{\hrule height 1pt}
        {\multirow{2}{*}{Methods}} & \multicolumn{4}{c|}{Performance} & \multicolumn{2}{c}{Adapter Complexity} \\\cline{2-7}
        & mAP & R-1 & R-5 & R-10 & FLOPs(G) & Params(M) \\
        \hline
        CLIP(L) + LoRA-r(64) & 54.5 & 55.3 & 68.1 & 76.1&0.01&0.10 \\
        CLIP(L) + LoRA-r(256) & 55.7 & 56.9 & 69.7 & 77.3&0.05&0.39 \\
        CLIP(L) + LoRA-r(384) & 47.2 & 47.1 & 62.2  &69.1 &0.08&0.60 \\
        CLIP(L) + BNA & 68.5 & 69.4 & 79.1 & 84.0&0.08&0.60 \\
        \rowcolor[gray]{0.92}
        \textbf{CLIP(L) + PFA} & \textbf{73.2} & \textbf{76.4} & \textbf{85.2} & \textbf{88.2}&0.30&2.36 \\
        \noalign{\hrule height 1pt}
    \end{tabular}}
    \vspace{-3mm}
    \caption{
      Comparison with different adapters.
      BNA denotes the bottleneck adapter.
      LoRA-r($\mathrm{x}$) denotes the LoRA with rank $\mathrm{x}$ and we set $\mathrm{x}$ to 64, 256 and 384 for comparison.
      }
    \label{tab:adapter_comparison}
    \vspace{-4mm}
  \end{table}
  \begin{figure}[tb]
    \centering
    \includegraphics[width=0.43\textwidth]{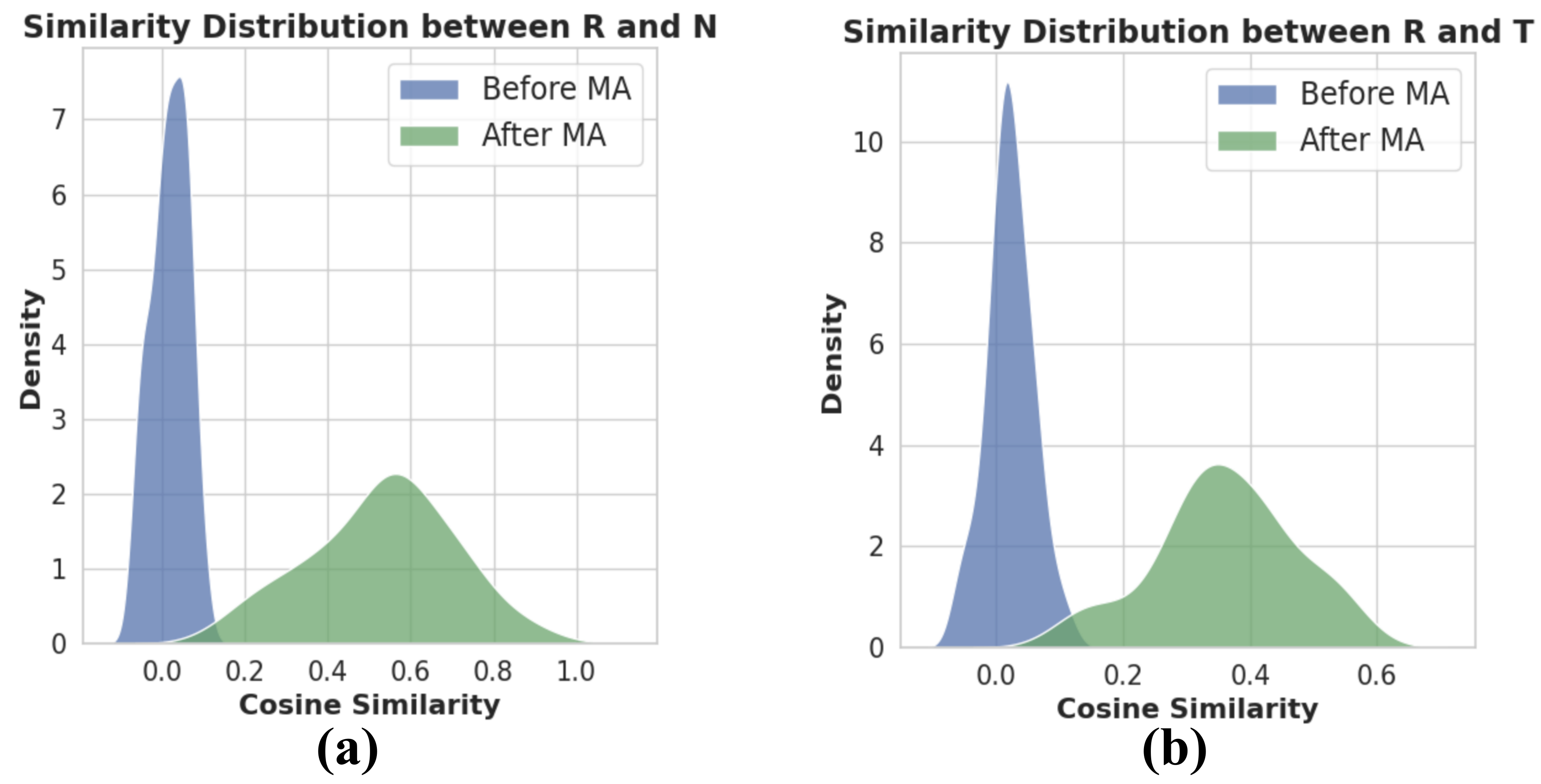}
    \vspace{-3mm}
    \caption{Alignment visualization of cosine similarity distributions across different modalities before and after MA.}
    \vspace{-2.5mm}
    \label{fig:similarity}
  \end{figure}
  \begin{table}[tb]
    \centering
    \renewcommand\arraystretch{1.15}
    \setlength\tabcolsep{1.5pt}
    \label{tab:method_comparison}
    \resizebox{0.45\textwidth}{!}
    {
    \begin{tabular}{c|cccc|cc}
      \noalign{\hrule height 1pt}
        {\multirow{2}{*}{Methods}} & \multicolumn{4}{c|}{Performance} & \multicolumn{2}{c}{Aggregation Complexity} \\\cline{2-7}
        & mAP & R-1 & R-5 & R-10 & FLOPs(G) & Params(M) \\
        \hline
        Sum & 75.3 & 79.4 & 87.2 & 89.4 &0.01&1.58\\
        Concat & 74.3 & 75.6 & 84.9 & 88.9&0.10&2.37 \\
        Transformer & 76.5 & 80.2 & 87.7 & 90.6&1.36&4.73\\
        Intra-MA & 76.2 & 80.9 & 88.2 & 90.5 &0.71&5.64 \\
        Inter-MA  & 77.5 & 81.2 & 88.6 & 90.9 &0.71&5.41\\
        \rowcolor[gray]{0.92}
        \textbf{MA} & \textbf{78.9} & \textbf{83.4} & \textbf{89.8} & \textbf{91.9} &1.42&9.47\\
        \noalign{\hrule height 1pt}
    \end{tabular}}
    \vspace{-3mm}
    \caption{Performance with different aggregation methods.}
    \label{tab:Main_fusion_comparison}
    \vspace{-5mm}
  \end{table}
  \subsection{Visualization}
  \textbf{Feature Distributions.}
  In Fig. \ref{fig:tsne}, we show the feature distributions with different modules.
  Comparing Fig. \ref{fig:tsne} (a) and Fig. \ref{fig:tsne} (b), CLIP(L) with PFA leads to more compact distributions across different IDs.
  Besides, with SRP, the gaps between different IDs in Fig. \ref{fig:tsne} (c) are further expanded.
  Finally, with MA, the feature distributions become more separable as shown in Fig. \ref{fig:tsne} (d).
  These visualizations strongly support the effectiveness of our proposed modules.
  \\
  \textbf{Feature Alignment of MA.}
  In Fig. \ref{fig:similarity}, we visualize the similarity distributions of different modalities before and after MA.
  The results clearly show that MA can fully align the distributions of different modalities, enhancing the feature aggregation.
  More details are in the supplementary material.
  \section{Conclusion}
  In this paper, we propose a novel feature learning framework named MambaPro, for multi-modal object ReID.
  We first employ a Parallel Feed-Forward Adapter (PFA) for knowledge transfer from the pre-trained CLIP to the ReID task.
  To guide joint learning of multi-modal features, we introduce the Synergistic Residual Prompt (SRP) for interactions of modality-specific prompts.
  Meanwhile, we propose a Mamba Aggregation (MA) to efficiently aggregate tokenized sequences from different modalities.
  With linear complexity, the MA efficiently handles long sequences, outperforming existing attention-based methods with low consumption.
  Extensive experiments on three multi-modal object ReID benchmarks validate the effectiveness of our method and show the potentials of fundamental models.
\section{Acknowledgments}
This work was supported in part by the National Natural Science Foundation of China (No.62101092, 62476044, 62388101), Open Project of Anhui Provincial Key Laboratory of Multimodal Cognitive Computation, Anhui University (No.MMC202102, MMC202407) and Fundamental Research Funds for the Central Universities (No.DUT23BK050, DUT23YG232).
\bibliography{aaai25}
\section{A. Introduction}
The supplementary material validates the effectiveness of our MambaPro with additional evidence.
We provide a comprehensive analysis of the model's performance, including the impact of various hyperparameters and the model's generalization to vehicle datasets.
To be specific, the supplementary material is composed of the following parts:
\begin{enumerate}
    \item \textbf{More method details:}
    \begin{itemize}
        \item Background knowledge of Mamba
        \item Workflow of our Synergistic Residual Prompt (SRP)
    \end{itemize}

    \item \textbf{More experimental details:}
    \begin{itemize}
        \item Implementation details
    \end{itemize}

    \item \textbf{Module validation and hyper-parameter analysis:}
    \begin{itemize}
        \item Model parameter comparison with other methods
        \item Analysis of different structures and hyper-parameters
        \item Generalization of the model on vehicle datasets
    \end{itemize}

    \item \textbf{Grad-CAM visualizations:}
    \begin{itemize}
        \item Visualizations for both person and vehicle ReID
    \end{itemize}
\end{enumerate}
These comprehensive experiments fully validate the effectiveness of our MambaPro.

\section{B. Methodology Details}
\subsection{Background knowledge of Mamba}
Mamba~\cite{gu2023mamba} is a sequential modeling framework designed to effectively capture complex interactions within ultra-long sequences by leveraging State Space Models (SSMs).
SSMs~\cite{gu2021efficiently,gu2021combining,smith2022simplified} are a type of sequence-to-sequence models inspired by continuous linear time-invariant (LTI) systems.
With linear complexity, they can effectively capture the dynamics inherent in the system's state variables, specifically defined as:
\begin{equation}
  y(t) = Ch(t) + Dx(t),
  \dot{h}(t) = Ah(t) + Bx(t).
\end{equation}
Here, $x(t) \in \mathbb{R}$ means the input, $h(t) \in \mathbb{R}^{N}$ denotes the hidden state and $y(t) \in \mathbb{R}$ represents the output.
$\dot{h}(t)$ stands for the time derivative of $h(t)$.
Besides, ${A} \in \mathbb{R}^{N \times N}$, ${B} \in \mathbb{R}^{N \times 1}$, ${C} \in \mathbb{R}^{1 \times N}$ and ${D} \in \mathbb{R}$ are the system matrices, where $N$ is the hidden state dimension.
Meanwhile, we often need to process discrete sequences like image and text.
Thus, SSMs utilize zero-order hold discretization to discretize the system matrices.
Then, the input sequence $x = \{x_1, x_2, ..., x_K\}$ can be mapped to the output sequence $y = \{y_1, y_2, ..., y_K\}$ with the following equations:
\begin{equation}
  y_k = \overline{C}{h_k} + \overline{D}{x_k},
  h_{k} = \overline{A}{h_{k-1}} +\overline{B}{x_k}.
\end{equation}
Although effective for sequence modeling, SSMs still faces limitations for the LTI property.
To address this issue, the Selective State Space Model (S6)~\cite{gu2023mamba} is introduced to make SSMs data-dependent.
With the selection mechanism, S6 can effectively model the complex interactions in ultra-long sequences with linear complexity.
In this work, we specifically design the intra-modality MA and inter-modality MA with S6, which can fully integrate complementary information from multiple modalities.
\begin{figure}[t]
  \centering
  \includegraphics[width=1.0\linewidth, height=0.37\linewidth]{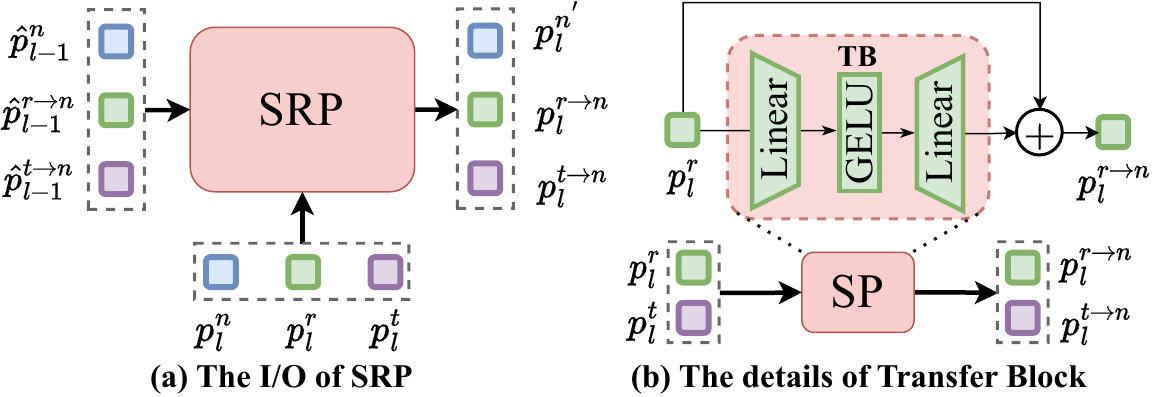}
   \caption{Detailed workflow of our SRP.}
   \label{fig:SRP}
\end{figure}
\begin{table*}[t]
  \centering
  \renewcommand\arraystretch{1.2}
  \setlength\tabcolsep{1.5pt}
  \resizebox{0.88\textwidth}{!}
  {
  \begin{tabular}{ccccccccc}
    \noalign{\hrule height 1pt}
    \multicolumn{1}{c}{\multirow{2}{*}{Methods}} &\multicolumn{1}{c}{\multirow{2}{*}{Params(M)}} &  \multicolumn{2}{c}{RGBNT201} &  \multicolumn{2}{c}{RGBNT100} & \multicolumn{2}{c}{MSVR310} \\
    \cline{3-8}
    & & mAP & Rank-1& mAP & Rank-1 & mAP & Rank-1 \\
    \hline
  MUDeep~\cite{qian2017multi} &77.75& 23.8 & 19.7 & - & - & - & - \\
  HACNN~\cite{li2018harmonious} &10.50& 21.3 & 19.0 & - & - & - & -\\
  MLFN~\cite{chang2018multi}&95.57 & 26.1 & 24.2 & - & - & - & - \\
  CAL~\cite{rao2021counterfactual} &97.62& 27.6 & 24.3 & - & - & - & - \\
  PCB~\cite{sun2018beyond}  &72.33& 32.8 & 28.1 & 57.2 &83.5 &23.2 &42.9 \\
  OSNet~\cite{zhou2019omni} &7.02& 25.4 & 22.3 & 75.0 &95.6 &28.7 &44.8 \\
  HAMNet~\cite{li2020multi} &  78.00 &27.7 &26.3 &74.5 &93.3 &27.1 &42.3\\
  CCNet~\cite{zheng2022multi} &  74.60 &- &- & 77.2 &96.3 &36.4 &\underline{55.2}\\
  IEEE~\cite{wang2022interact} & 109.22 &49.5&48.4 &-&-&-&-\\
  GAFNet~\cite{guo2022generative} &  130.00 &- &- &74.4 &93.4 &- &-\\
  LRMM~\cite{wu2025lrmm} &  86.40 &52.3 &51.1 &78.6 &\textbf{96.7} &36.7 &49.7\\
  \hline
  TransReID$^*$~\cite{he2021transreid} & 278.23 &- &- &75.6 &92.9 &18.4&29.6 \\
  UniCat$^*$~\cite{crawford2023unicat}  & 259.02 &57.0 &55.7&79.4&96.2&- &- \\
  GraFT$^*$~\cite{yin2023graft} & 101.00 &- &- &76.6&94.3&-&-\\
  TOP-ReID$^*$~\cite{wang2023top} & 324.53 &\underline{72.3} &\underline{76.6} &81.2&96.4&35.9&44.6\\
  EDITOR$^*$~\cite{zhang2024magic} &118.55 & 66.5       & 68.3& 82.1 & {96.4} &39.0 & 49.3\\
  RSCNet$^*$~\cite{yu2024representation} &  124.10 & 68.2 & 72.5 &\underline{82.3} &\underline{96.6} &\underline{39.5} &49.6\\
  \hline
  \rowcolor[gray]{0.92}
  MambaPro$\ddagger$ & 74.20 &\textbf{78.9}&\textbf{83.4}&\textbf{83.9} &94.7 & \textbf{47.0} &\textbf{56.5}\\
  \noalign{\hrule height 1pt}
  \end{tabular}
  }
  \caption{Parameter comparison  in our framework.
  The top results are highlighted in bold, with the second-best underlined.
  Symbols: $\ddagger$ indicates CLIP-based methods, $*$ marks ViT-based methods and others are CNN-based methods.}
  \label{tab:params}
  \end{table*}
\subsection{Workflow of the SRP}
As shown in Fig.~\ref{fig:SRP} (a), we take the NIR modality as an example to clarify it.
In fact, SRP's input consists of two parts: (1) prompt tokens $\hat{p}_{l-1}^{n},\hat{p}_{l-1}^{r \to n},\hat{p}_{l-1}^{t \to n}$ from the output of the last Transformer layer; (2) prompt tokens $p_{l}^{n}, p_{l}^{r}, p_{l}^{t}$ to be added in the current layer.
The output includes $p_{l}^{n^{'}}$ from RP and ${p}_{l}^{r \to n}$, ${p}_{l}^{t \to n}$ from SP.
As shown in Fig.~\ref{fig:SRP} (b), SP utilizes the Transfer Block (TB) to transfer complementary information from prompts of other modalities.
Each modality utilizes its specific TB.
RP aggregates information from $\hat{p}_{l-1}^{n},\hat{p}_{l-1}^{r \to n},\hat{p}_{l-1}^{t \to n}$ and injects the aggregated information into $p_{l}^{n}$, forming $p_{l}^{n^{'}}$ with different levels of information.
Finally, we concatenate the output of SRP with the class token and patch tokens, as the input of the current Transformer layer for further processing.
\section{C. Experimental Details}
\subsection{Implementation Details}
For the input resolution, images are resized to 256$\times$128 for RGBNT201 and 128$\times$256 for RGBNT100/MSVR310.
For data augmentation, we employ random horizontal flipping, cropping and erasing~\cite{zhong2020random}.
A warmup strategy with a cosine decay is used for the learning rate scheduling.
On the three datasets, we use a learning rate of 3.5$\mathrm{e}^{-4}$ for PEFT.
To compare with the fully fine-tuned CLIP, we use a learning rate of 5$\mathrm{e}^{-6}$ on the CLIP backbone, while other parts are trained with 3.5$\mathrm{e}^{-4}$.
In the small-scale datasets RGBNT201 and MSVR310, the mini-batch size is set to 64, 4 images sampled for each identity and 16 identities sampled for each batch.
Both of the two datasets are trained with 60 epochs and the warm up epoch is set to 10.
For the large-scale RGBNT100 dataset, to better utilize the data, we set the mini-batch size to 128, with 16 images sampled for each identity.
This leads to a more stable training.
Besides, we train the model for 30 epochs with a 5-epoch warm-up.
\begin{figure*}[t]
  \centering
  \includegraphics[width=0.9\textwidth]{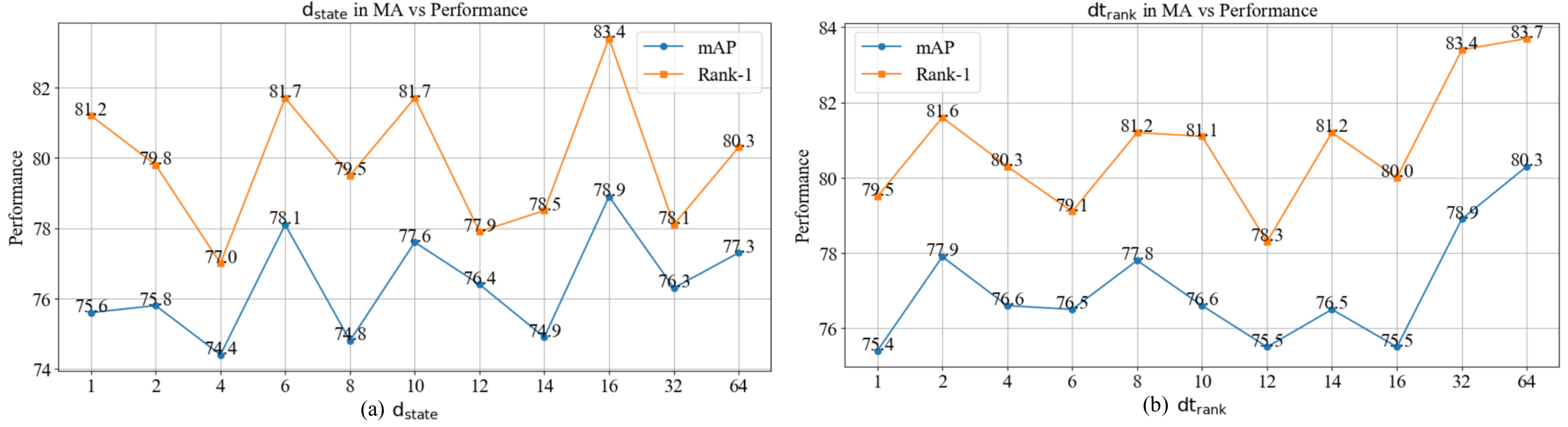}
  \caption{
    Performance comparison with different $\mathrm{d_{state}}$ and $\mathrm{dt_{rank}}$ in MA.
  }
  \label{fig:state-rank}
\end{figure*}
\begin{figure*}[t]
  \centering
  \includegraphics[width=0.9\textwidth]{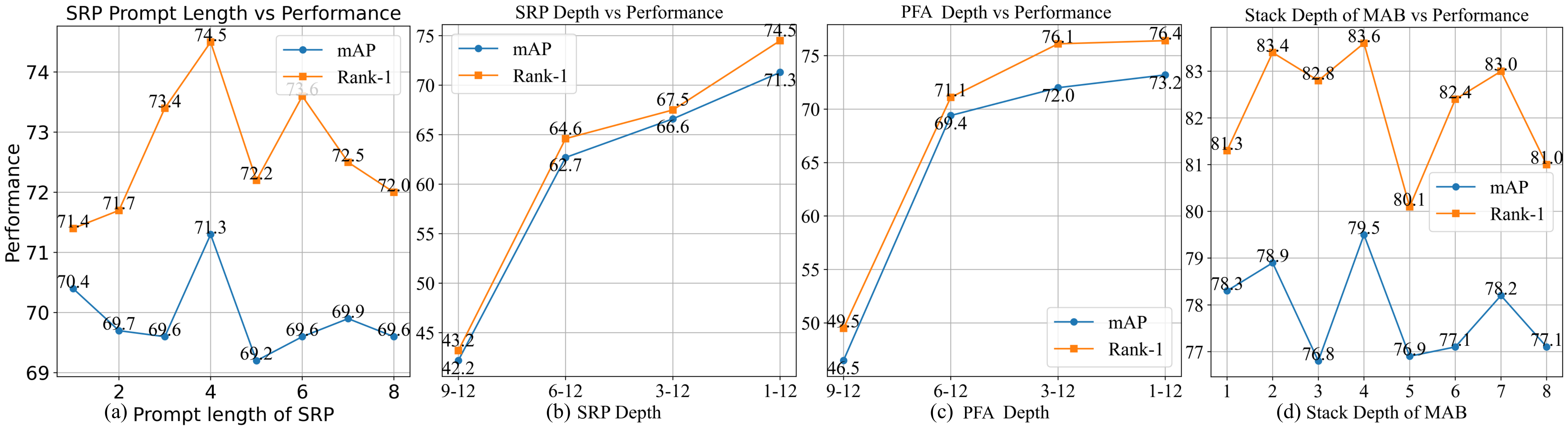}
  \caption{
    Comparison of different depths or prompt lengths in proposed modules.
  }
  \label{fig:depth}
  \vspace{2mm}
\end{figure*}
\section{D. Experimental Analysis}
\subsection{Parameter Comparisons}
In Tab. \ref{tab:params}, we compare the trainable parameters of our model with other methods, including both CNN-based methods~\cite{zhou2019omni,sun2018beyond,rao2021counterfactual,chang2018multi,li2018harmonious,qian2017multi,li2020multi,wang2022interact,zheng2022multi,guo2022generative} and Transformer-based methods~\cite{he2021transreid,wang2023top,crawford2023unicat,yin2023graft}.
In general, CNN-based methods have fewer parameters than Transformer-based methods.
However, in Transformer-based methods, our MambaPro achieves the best performance with the fewest trainable parameters.
In contrast, the parameter count of the other Transformer-based methods is at least 100M, with the highest reaching 324.53M.
With comparable parameters as CNN-based methods, our MambaPro outperforms them by a large margin, verifying the effectiveness of our method.
\subsection{Different Hyper-parameters and Structures}
\textbf{Effect of $\mathrm{d_{state}}$ and $\mathrm{dt_{rank}}$ in MA.}
In Fig. \ref{fig:state-rank}, we compare the performance with different $\mathrm{d_{state}}$ and $\mathrm{dt_{rank}}$ in MA.
In fact, $\mathrm{d_{state}}$ is the dimension of the hidden state $h_{k}$, while $\mathrm{dt_{rank}}$ is the rank when generating the discrete time steps $\mathrm{\Delta}$ in SSM.
As shown in Fig. \ref{fig:state-rank} (a), the performance first rises and then falls with the increase of $\mathrm{d_{state}}$.
The best performance is achieved when $\mathrm{d_{state}}$ is set to 16.
The reason is that a small $\mathrm{d_{state}}$ is not able to capture the complex multi-modal correlations, while a large $\mathrm{d_{state}}$ may lead to overfitting.
As for $\mathrm{dt_{rank}}$, the performance is relatively stable.
However, the performance exhibits an evident increase when $\mathrm{dt_{rank}}$ is bigger than 32.
The bigger $\mathrm{dt_{rank}}$ can avoid information loss, which is beneficial for the performance improvement.
\\
\textbf{Effect of Varying Depths of Proposed Modules.}
In Fig. \ref{fig:depth}, we show the performance with different depths of PFA, SRP and MA.
Meanwhile, we also perform the analysis on the length of prompts.
Fig. \ref{fig:depth} (a) shows that with more learnable prompts in each modality, the performance first increases and then decreases, achieving the best performance with $N_{pr}=4$.
In Fig. \ref{fig:depth} (b), the performance shows a large improvement with the depth increasing from 9-12 to 6-12.
Furthermore, the best performance is achieved with SRP in every layer.
Similar trends can be observed in Fig. \ref{fig:depth} (c), with the best performance achieved with all layers.
Meanwhile, the overall performance is higher than that of Fig. \ref{fig:depth} (b), indicating the strong transfer capability of PFA.
For Fig. \ref{fig:depth} (d), stacked MA blocks improve the performance, with a 79.5\% mAP achieved with 4 MA blocks.
Considering the complexity, we choose 2 MA blocks as the best trade-off.
These results verify the effectiveness of our proposed modules.
\\
\textbf{Effect of the Inserted Position of PFA.}
In Tab. \ref{tab:adapter_position}, we compare the performance with different inserted positions of PFA.
Within a Transformer layer, the main components include the Multi-Head Self-Attention (MHSA) and the Feed-Forward Network (FFN).
Thus, we insert the PFA behind the FFN and the MHSA or on the side of the FFN and the MHSA, respectively.
We observe that the PFA inserted on the side of original components achieves better performance.
The main reason is that the stable structure of original components is not disturbed.
By introducing additional knowledge from special domains with our proposed PFA, the pre-trained knowledge can be better utilized.
\begin{table}[t]
  \centering
  \renewcommand\arraystretch{1.2}
  \setlength\tabcolsep{1.5pt}
  \resizebox{0.35\textwidth}{!}
  {
  \begin{tabular}{ccccc}
    \noalign{\hrule height 1pt}
      {\multirow{2}{*}{Methods}} & \multicolumn{4}{c}{RGBNT201} \\\cline{2-5}
      & mAP & Rank-1 & Rank-5 & Rank-10  \\
      \hline
      \rowcolor[gray]{0.92}
      Side (FFN) & \textbf{73.2} & \textbf{76.4} & \textbf{85.2} & \textbf{88.2}\\
      Behind (FFN) & 46.9 & 46.2 & 63.4 & 72.7 \\
      Side (MHSA) &  \underline{66.3} & \underline{67.2} & \underline{79.7}  &\underline{84.2}  \\
      Behind (MHSA) & 52.0 & 52.3 & 66.4 & 72.8 \\
      \noalign{\hrule height 1pt}
  \end{tabular}}
  \caption{Comparison with different positions of PFA.}
  \label{tab:adapter_position}
\end{table}
\begin{table}[t]
  \centering
  \renewcommand\arraystretch{1.2}
  \setlength\tabcolsep{1.5pt}
  \resizebox{0.30\textwidth}{!}
  {
  \begin{tabular}{ccccc}
    \noalign{\hrule height 1pt}
      {\multirow{2}{*}{Methods}} & \multicolumn{4}{c}{RGBNT201} \\\cline{2-5}
      & mAP & Rank-1 & Rank-5 & Rank-10  \\
      \hline
      ViT (L) &  67.5 & 71.3 & 82.1  &88.6  \\
      ViT (F) &  67.6 & 70.2 & 83.1  &87.4  \\
      \rowcolor[gray]{0.92}
      CLIP (L) & \textbf{78.9} & \underline{83.4} & \textbf{89.8} & \underline{91.9}\\
      CLIP (F) & \underline{78.7} &\textbf{83.5} & \underline{89.6} & \textbf{92.9}\\
      \noalign{\hrule height 1pt}
  \end{tabular}}
  \caption{Comparison with different pre-trained backbones.}
  \label{tab:pretrained_backbone}
\end{table}
\begin{figure*}[t]
  \centering
  \includegraphics[width=0.8\textwidth]{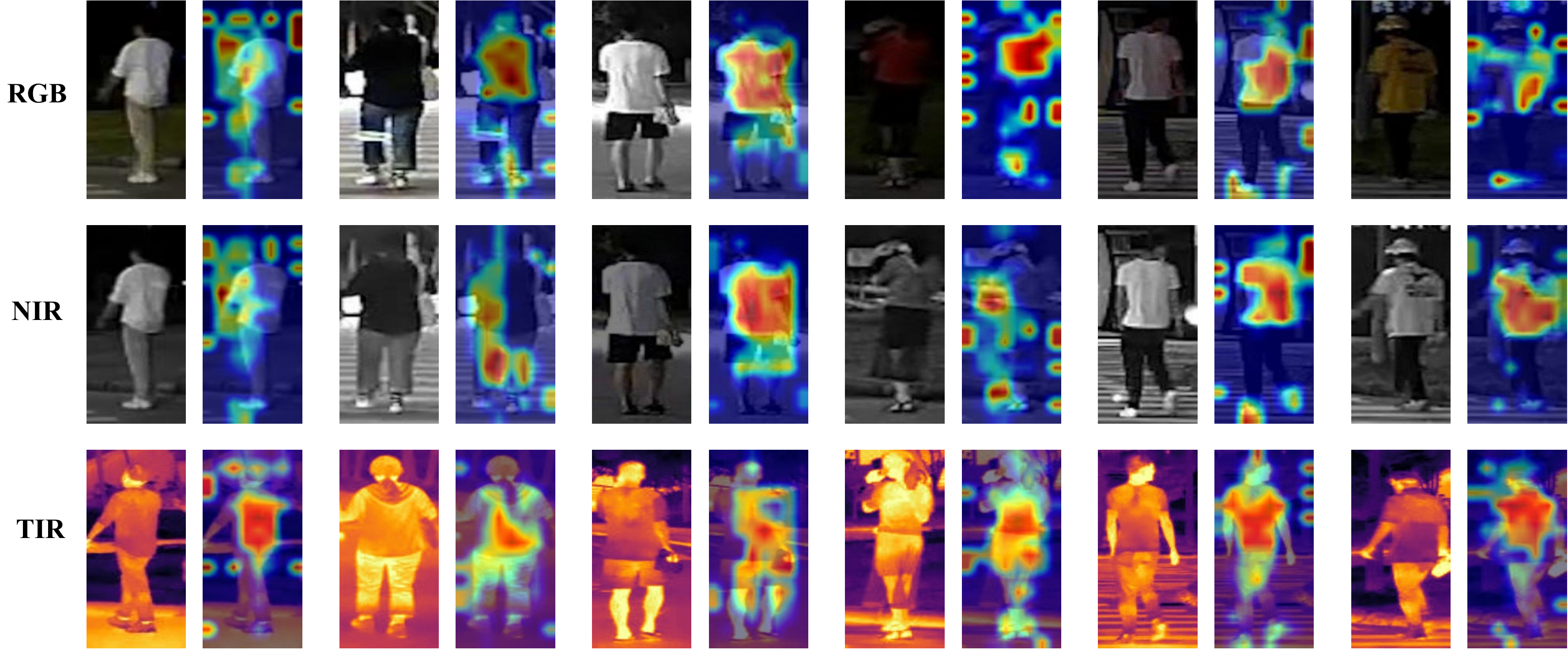}
  \caption{
    Grad-CAM visualizations of multi-modal person ReID.
  }
  \label{fig:person-gradcam}
\end{figure*}
\begin{figure*}[t]
  \centering
  \includegraphics[width=0.8\textwidth]{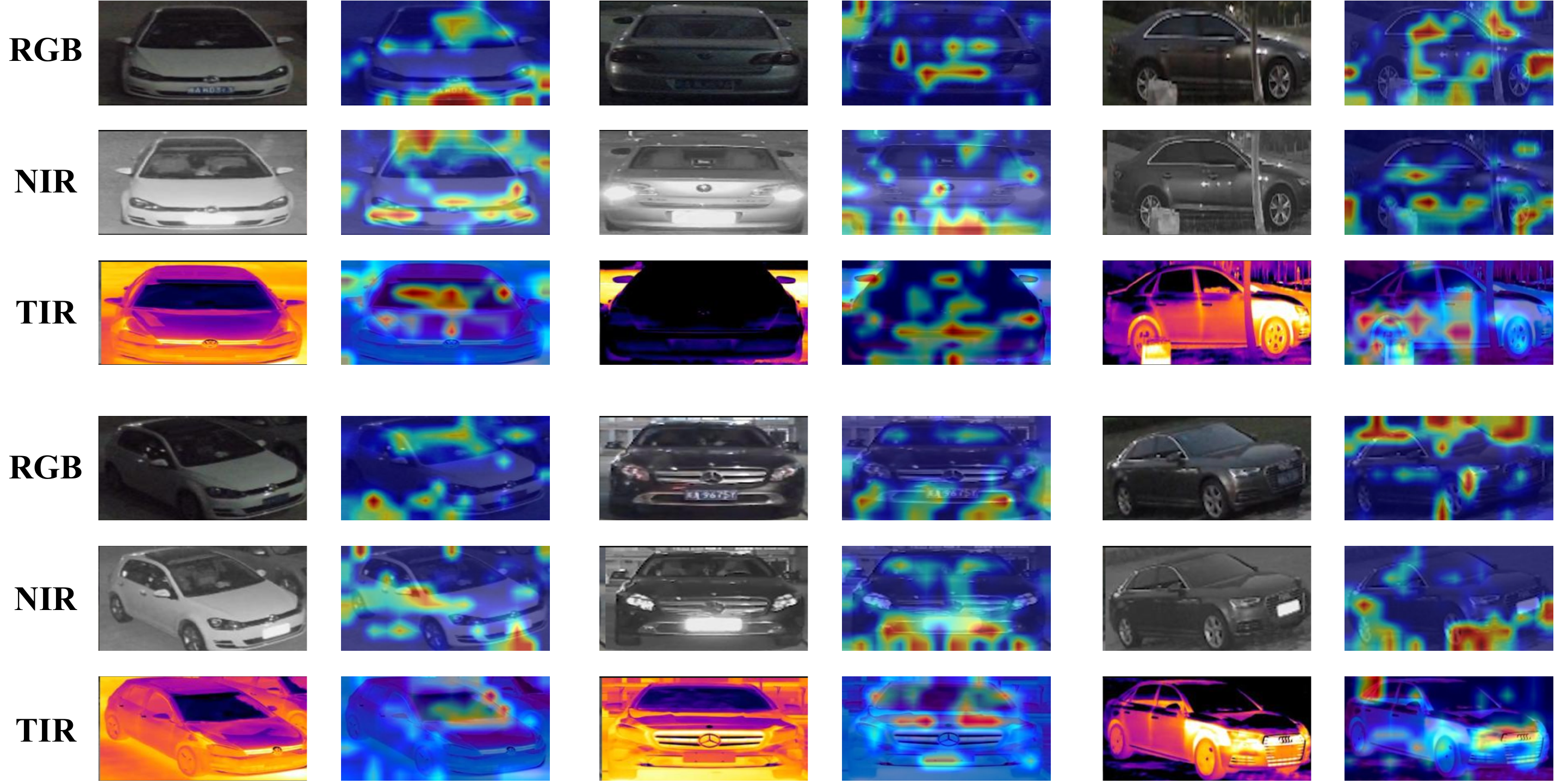}
  \caption{
    Grad-CAM visualizations of multi-modal vehicle ReID.
  }
  \label{fig:vehicle-gradcam}
\end{figure*}
\\
\textbf{Effect of the Pre-trained Backbone.}
In fact, our framework can also change the backbone to ViT, which is pre-trained on the ImageNet.
Thus, we show the performance with different pre-trained backbones in Tab. \ref{tab:pretrained_backbone}.
Comparing the first line and the third line, we observe that CLIP (L) achieves better performance than ViT (L), which indicates the effectiveness of the CLIP pre-trained knowledge in multi-modal object ReID.
Meanwhile, ViT (F) achieves competitive performance, showcasing the effectiveness of our modules.
\\
\textbf{Comparison with CLIP-based TOP-ReID.}
In Tab.~\ref{tab:pre}, we compare our MambaPro with the CLIP-based TOP-ReID, with both models \textbf{fully fine-tuned}.
The results show that TOP-ReID with CLIP underperforms compared to our MambaPro.
Despite having nearly half the parameters, MambaPro achieves a 5.4\% improvement in mAP and a 6.9\% improvement in Rank-1.
These findings suggest that our MambaPro can better leverage CLIP's knowledge.
\begin{table}[t]
  \vspace{-0mm}
  \centering
  \renewcommand\arraystretch{1.2}
  \setlength\tabcolsep{1 pt}
  \resizebox{0.47\textwidth}{!}
  {
  \begin{tabular}{ccccccc}
    \noalign{\hrule height 1pt}
  {\multirow{2}{*}{Methods}} &{\multirow{2}{*}{Params(M)}}&{\multirow{2}{*}{Pretrained}}&  \multicolumn{4}{c}{RGBNT201} \\\cline{4-7}
  &&&mAP&Rank-1 &Rank-5 &Rank-10 \\ \hline
  TOP-ReID & \multirow{1}{*}{324.53}&  CLIP &\underline{73.3} &\underline{77.2} &\underline{85.9} &\underline{90.1}  \\
  \rowcolor[gray]{0.92}
  MambaPro &\multirow{1}{*}{160.34} & CLIP &\textbf{78.7} &\textbf{83.5} &\textbf{89.6} &\textbf{92.9}  \\
  \noalign{\hrule height 1pt}
  \end{tabular}
  }
  \caption{Comparison with CLIP-based TOP-ReID.}
  \label{tab:pre}
\end{table}
\begin{table}[t]
  \centering
  \renewcommand\arraystretch{1.15}
  \setlength\tabcolsep{1.5pt}
  \label{tab:method_comparison}
  \resizebox{0.45\textwidth}{!}
  {
  \begin{tabular}{c|cccc|cc}
    \noalign{\hrule height 1pt}
      {\multirow{2}{*}{Methods}} & \multicolumn{4}{c|}{Performance} & \multicolumn{2}{c}{Aggregation Complexity} \\\cline{2-7}
      & mAP & R-1 & R-5 & R-10 & FLOPs(G) & Params(M) \\
      \hline
      MA(Trans) & 77.0 & 80.4 & 88.4 & 90.9&1.81&14.19\\
      \rowcolor[gray]{0.92}
      \textbf{MA(Mamba)} & \textbf{78.9} & \textbf{83.4} & \textbf{89.8} & \textbf{91.9} &1.42&9.47\\
      \noalign{\hrule height 1pt}
  \end{tabular}}
  \vspace{-0mm}
  \caption{Performance with different aggregation methods.}
  \label{tab:fusion_comparison}
  \vspace{-0mm}
\end{table}
\\
\textbf{Effect of Mamba Block in MA.}
To clarify, the complete MA module includes both intra-modal and inter-modal interactions.
A fair comparison is shown in Tab. 6 between Row 3 (standard Transformer applied across all three modalities) and Row 5 (using Mamba blocks).
Mamba achieves a 1\% mAP increase with nearly half the FLOPs of the Transformer.
To further enhance modeling capability, we add Intra-MA, and the final MA achieves a 3.4\% mAP improvement with FLOPs and parameters comparable to the Transformer block.
Additionally, in Tab.~\ref{tab:fusion_comparison}, we validate MA's effectiveness by replacing Mamba blocks in both MA-inter and MA-intra with Transformers, denoted as MA(Trans), verifying the superiority of our proposed Mamba-based MA.
\begin{table}[t]
  \centering
  \renewcommand\arraystretch{1.2}
  \setlength\tabcolsep{2pt}
  \resizebox{0.474\textwidth}{!}
  {
  \begin{tabular}{c|ccccc|cccc|cc}
    \noalign{\hrule height 1pt}
      & \multicolumn{5}{c|}{Methods} & \multicolumn{4}{c|}{RGBNT100} & \multicolumn{2}{c}{Complexity} \\
        \cline{2-12}
        &CLIP(L)&CLIP(F)&PFA & SRP &MA & mAP & Rank-1 & Rank-5 & Rank-10 & Mem.(G) & Param.(M) \\
      \hline
      A&\ding{51}&\ding{53}&\ding{53}&\ding{53}&\ding{53}& 17.9 & 40.9 & 50.5 & 56.0 &21.79 &0 \\
      B&\ding{53}&\ding{51}&\ding{53}&\ding{53}&\ding{53}& 80.7 & 93.5 & \textbf{96.2} & \textbf{96.7} &28.73 &86.14 \\
      C&\ding{51}&\ding{53}&\ding{51}&\ding{53}&\ding{53}& 81.6 & 93.9 & 95.0 & 95.6 &28.21 &28.34\\
      D&\ding{51}&\ding{53}&\ding{53}&\ding{51}&\ding{53}& 79.1 & 92.9 & 94.2 & 94.9 &24.65 &28.48 \\
      E&\ding{51}&\ding{53}&\ding{51}&\ding{51}&\ding{53}& 82.0 & 94.0 & \underline{95.3} & \underline{95.8} &32.05 &56.82 \\
      \rowcolor[gray]{0.92}
      \textbf{F}&\ding{51}&\ding{53}&\ding{51}&\ding{51}&\ding{51}& \textbf{83.9} & \textbf{94.7} & 94.9 & 95.3 &38.01&74.20 \\
      G&\ding{53}&\ding{51}&\ding{51}&\ding{51}&\ding{51}& \underline{83.1} & \underline{94.1} & 94.9 & 95.6 &45.36 &160.34 \\
      \noalign{\hrule height 1pt}
  \end{tabular}
  }
  \caption{Performance comparison with key components.}
  \label{tab:ablation_vehicle}
\end{table}
\subsection{Generalization to Vehicle Datasets}
In Tab. \ref{tab:ablation_vehicle}, we validate the effectiveness of key components on the vehicle dataset RGBNT100.
Here, the memory consumption is measured with a batch size of 128.
Comparing Model A with Model B, we observe a remarkable improvement in mAP, showcasing the necessity of knowledge transfer.
In Model C, PFA outperforms the fully fine-tuned CLIP, showcasing the effectiveness of parallel knowledge transfer.
Meanwhile, Model D shows competitive performance, showing the effectiveness of SRP.
Furthermore, through the combination of SRP and PFA, Model E achieves higher performance.
Finally, with the MA module, Model F achieves the best performance.
Similarly, we also compare our method with the fully fine-tuned CLIP, which has a much higher complexity.
In contrast, our method achieves better performance with lower complexity, validating the effectiveness of our proposed method on vehicle datasets.
\section{E. Visualization Results}
In this part, we provide the Grad-CAM visualizations of $f^{M}_{ir}, M \in \{n,r,t\}$.
These features are the fully interacted features of different modalities after MA.
To be specific, we visualize discriminative features of persons and vehicles in Fig. \ref{fig:person-gradcam} and Fig. \ref{fig:vehicle-gradcam}, respectively.
As we can see, different modalities have different discriminative regions, which means the complementary information is effectively utilized.
Meanwhile, in Fig. \ref{fig:person-gradcam}, the discriminative regions are consistent with the human perception, where fine-grained details like the head, clothes, shoes and bags are highlighted.
In Fig. \ref{fig:vehicle-gradcam}, we exhibit the samples with different visual scenarios and viewpoints.
We observe that even certain modalities are completely invisible, the discriminative regions are still highlighted, showcasing the effectiveness of modality interaction.
In conclusion, the Grad-CAM visualizations fully validate the effectiveness of our proposed method.
\end{document}